\documentclass[10pt,twocolumn,letterpaper]{article}

\usepackage[pagenumbers]{cvpr} % To force page numbers, e.g. for an arXiv version

\usepackage[dvipsnames]{xcolor}

\definecolor{myPurple}{rgb}{0.4, .0, .8}
\definecolor{myGreen}{rgb}{0, .8, .3}
\definecolor{myRed}{rgb}{0.8, .2, .2}
\definecolor{myPink}{rgb}{0.8, .5, .5}
\definecolor{myOrange}{rgb}{0.7, 0.45, 0.2}
\definecolor{myBlue}{rgb}{.0, .0, 1.0}
\definecolor{myBlue2}{rgb}{.0, .0, 0.5}
\definecolor{myBlack}{rgb}{.0, .0, 0.0}
\definecolor{mycyan}{rgb}{.39,.58,.93}

\newcommand{\ourname}{\color{myBlack}{GenN2N}}

\usepackage{url}
\usepackage{enumitem}
\usepackage{comment}
\usepackage{times}% added by kunming
\usepackage{multirow}
\usepackage{multicol}
\usepackage{booktabs}
\usepackage{bbding} % add Checkmark
\usepackage{float}
\usepackage{subcaption}
\usepackage{tabularx}
\usepackage{array}
\usepackage{mathtools}
\usepackage[title]{appendix}

\definecolor{cvprblue}{rgb}{0.21,0.49,0.74}
\usepackage[pagebackref,breaklinks,colorlinks,citecolor=cvprblue]{hyperref}
\usepackage[accsupp]{axessibility}

\def\paperID{11231} % *** Enter the Paper ID here
\def\confName{CVPR}
\def\confYear{2024}

\title{GenN2N: Generative NeRF2NeRF Translation}

\author{
Xiangyue Liu\textsuperscript{\rm 1} \ \ \ Han Xue\textsuperscript{\rm 2} \ \ \ Kunming Luo\textsuperscript{\rm 1} \ \ \ Ping Tan\textsuperscript{\rm 1}\footnotemark[2] \ \ \ Li Yi\textsuperscript{\rm 2,}\textsuperscript{\rm 3,}\textsuperscript{\rm 4}\footnotemark[2]\\
\\
\textsuperscript{\rm 1} Hong Kong University of Science and Technology \ \ \ 
\textsuperscript{\rm 2} Tsinghua University \\
\textsuperscript{\rm 3} Shanghai Artificial Intelligence Laboratory \ \ \
\textsuperscript{\rm 4} Shanghai Qi Zhi Institute
}

\begin{document}

\twocolumn[{%
\renewcommand\twocolumn[1][]{#1}%
\maketitle
\begin{center}
    \centering
    \captionsetup{type=figure}
    \includegraphics[width=0.85\linewidth]{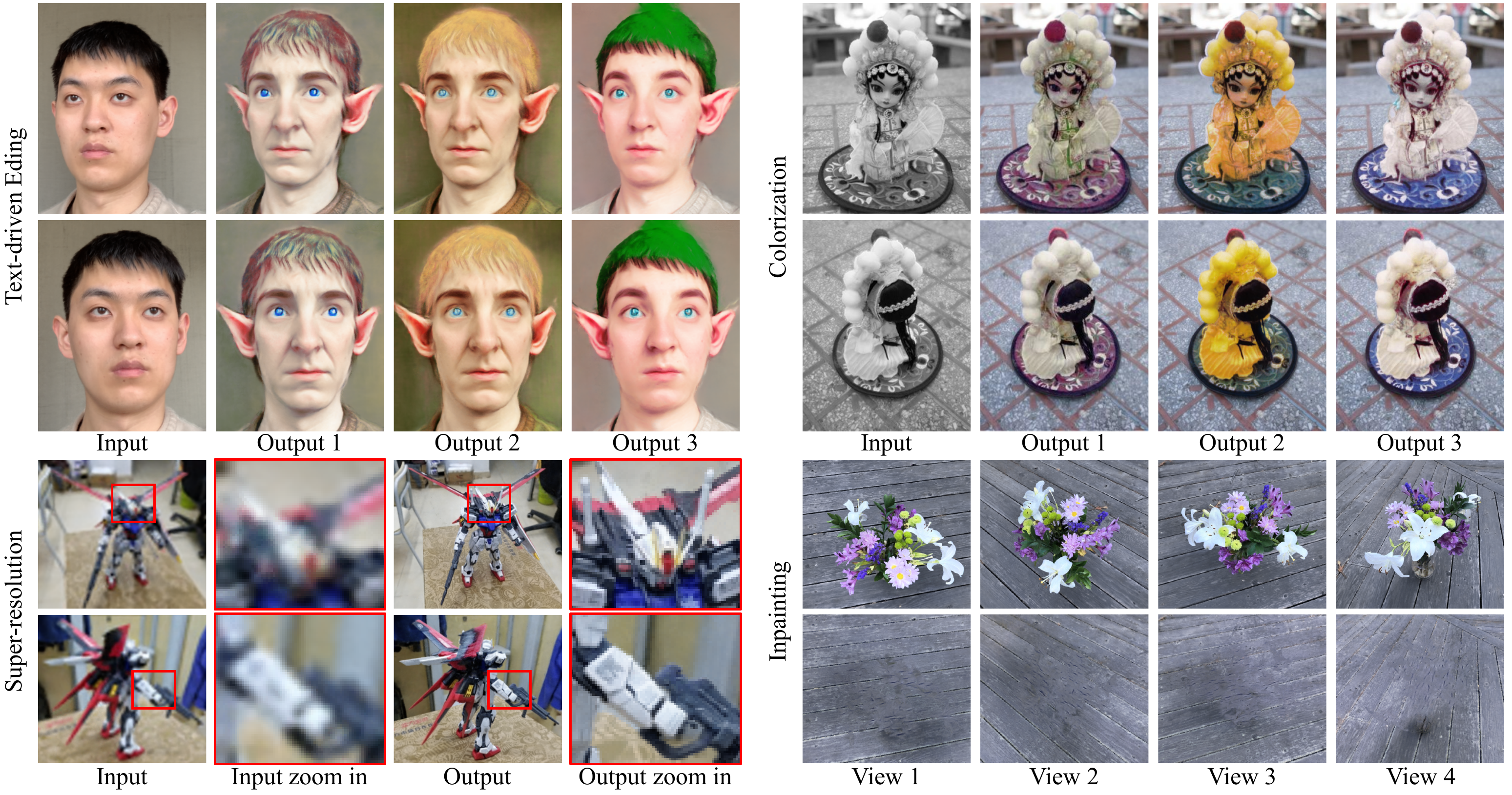}
    \vspace{-0.1in}
    \captionof{figure}{We introduce GenN2N, a unified framework for NeRF-to-NeRF translation, enabling a range of 3D NeRF editing tasks, including text-driven editing, colorization, super-resolution, inpainting, etc. We show at least two rendering views of edited NeRF scenes at inference time. Given a 3D NeRF scene, GenN2N can produce high-quality editing results with suitable multi-view consistency.}
    \label{fig:teaser}
\end{center}%
}]

\maketitle

\renewcommand{\thefootnote}{\fnsymbol{footnote}} %将脚注符号设置为fnsymbol类型，即特殊符号表示
% \footnotetext[1]{These authors contributed equally to this work.} %对应脚注[1]
\footnotetext[2]{Corresponding authors.} %对应脚注[2]

\begin{abstract}
We present GenN2N, a unified NeRF-to-NeRF translation framework for various NeRF translation tasks such as text-driven NeRF editing, colorization, super-resolution, inpainting, etc. Unlike previous methods designed for individual translation tasks with task-specific schemes, GenN2N achieves all these NeRF editing tasks by employing a plug-and-play image-to-image translator to perform editing in the 2D domain and lifting 2D edits into the 3D NeRF space. Since the 3D consistency of 2D edits may not be assured, we propose to model the distribution of the underlying 3D edits through a generative model that can cover all possible edited NeRFs. To model the distribution of 3D edited NeRFs from 2D edited images, we carefully design a VAE-GAN that encodes images while decoding NeRFs. The latent space is trained to align with a Gaussian distribution and the NeRFs are supervised through an adversarial loss on its renderings. To ensure the latent code does not depend on 2D viewpoints but truly reflects the 3D edits, we also regularize the latent code through a contrastive learning scheme. Extensive experiments on various editing tasks show GenN2N, as a universal framework, performs as well or better than task-specific specialists while possessing flexible generative power. More results on our project page:  \url{https://xiangyueliu.github.io/GenN2N/}.
\end{abstract}

\section{Introduction}
\label{sec:intro}

Over the past few years, Neural radiance fields (NeRFs)~\citep{mildenhall2021nerf} have brought a promising paradigm in the realm of 3D reconstruction, 3D generation, and novel view synthesis due to their unparalleled compactness, high quality, and versatility. Extensive research efforts have been devoted to creating NeRF scenes from 2D images~\citep{melas2023realfusion,yu2021pixelnerf,cai2022pix2nerf,wang2022rodin,liu2023zero} or just text~\citep{poole2022dreamfusion,jain2022zero} input. However, once the NeRF scenes have been created, these methods often lack further control over the generated geometry and appearance. NeRF editing has therefore become a notable research focus recently. 

Existing NeRF editing schemes are usually task-specific. For example, researchers have developed NeRF-SR~\citep{wang2022nerf}, NeRF-In~\citep{liu2022nerf}, PaletteNeRF~\citep{kuang2023palettenerf} for NeRF super-resolution, inpainting, and color-editing respectively. These designs require a significant amount of domain knowledge for each specific task. On the other hand, in the field of 2D image editing, a growing trend is to develop universal image-to-image translation methods to support versatile image editing~\citep{parmar2023zero, zhang2023adding, saharia2022palette}. By leveraging foundational 2D generative models, e.g., stable diffusion~\citep{rombach2022high}, these methods achieve impressive editing results without task-specific customization or tuning. We then ask the question: can we conduct universal NeRF editing leveraging foundational 2D generative models as well?

The first challenge is the representation gap between NeRFs and 2D images. It is not intuitive how to leverage image editing tools to edit NeRFs. A recent text-driven NeRF editing method~\citep{haque2023instruct} has shed some light on this. The method adopts a ``render-edit-aggregate'' pipeline. Specifically, it gradually updates a NeRF scene by iteratively rendering multi-view images, conducting text-driven visual editing on these images, and finally aggregating the edits in the NeRF scene. It seems that replacing the image editing tool with a universal image-to-image translator could lead to a universal NeRF editing method. However, the second challenge would then come. Image-to-image translators usually generate diverse and inconsistent edits for different views, e.g. turning a man into an elf might or might not put a hat on his head, making edits aggregation intricate. Regarding this challenge, Instruct-NeRF2NeRF~\citep{haque2023instruct} presents a complex optimization technique to pursue unblurred NeRF with inconsistent multi-view edits. Due to its complexity, the optimization cannot ensure the robustness of the outcomes. Additionally, the unique optimization outcome fails to reflect the stochastic nature of NeRF editing. Users typically anticipate a variety of edited NeRFs just like the diverse edited images.

To tackle the challenges above, we propose GenN2N, a unified NeRF-to-NeRF translation framework for various NeRF editing tasks such as text-driven editing, colorization, super-resolution, inpainting (see Fig.~\ref{fig:teaser}). In contrast to Instruct-NeRF2NeRF which adopts a ``render-edit-aggregate'' pipeline, we first render a NeRF scene into multi-view images, then exploit an image-to-image translator to edit different views, and finally learn a generative model to depict the distribution of NeRF edits. Instead of aggregating all the image edits to form a single NeRF edit, our key idea is to embrace the stochastic nature of content editing by modeling the distribution of the edits in the 3D NeRF space.

Specifically given a NeRF model or its multi-view images, along with the editing goal, we first generate edited multi-view images using a plug-and-play image-to-image translator. Each view corresponds to a unique 3D edit with some geometry or appearance variations. Conditioned on the input NeRF, GenN2N trains a conditional 3D generative model to reflect such content variations. At the core of GenN2N, we design a 3D VAE-GAN that incorporates a differentiable volume renderer to connect 3D content creation with 2D GAN losses, ensuring that the inconsistent multi-view renderings can still help each other regarding 3D generation. Moreover, we introduce a contrastive learning loss to ensure that the 3D content variation can be successfully understood just from edited 2D images without being influenced by the camera viewpoints. 
During inference, users can simply sample from the conditional generative model to obtain various 3D editing results aligned with the editing goal.
% items, and 360-degree environment
We have conducted experiments on human, items, indoor and outdoor scenes for various editing tasks such as text-driven editing, colorization, super-resolution and inpainting, demonstrating the effectiveness of GenN2N in supporting diverse NeRF editing tasks while keeping the multi-view consistency of the edited NeRF. 

We summarize the contribution of this paper as follows,
\begin{itemize}
\item A generative NeRF-to-NeRF translation formulation for the universal NeRF editing task together with a generic solution;
\item a 3D VAE-GAN framework that can learn the distribution of all possible 3D NeRF edits corresponding to the a set of input edited 2D images;
\item a contrastive learning framework that can disentangle the 3D edits and 2D camera views from edited images;
\item extensive experiments demonstrating the superior efficiency, quality, and diversity of the NeRF-to-NeRF translation results.
\end{itemize}

\section{Related Work}
\label{gen_inst}

\noindent{\bf NeRF Editing.} Previous works such as EditNeRF~\citep{liu2021editing} propose a conditional neural field that enables shape and appearance editing in the latent space. PaletteNeRF~\citep{kuang2023palettenerf,wu2022palettenerf} focuses on controlling color palette weights to manipulate appearance. Other approaches utilize bounding boxes~\citep{zhang2021editable}, meshes~\citep{yuan2022NeRF}, point clouds~\citep{chen2023neuraleditor}, key points~\citep{zheng2022editableNeRF}, or feature volumes~\citep{lazova2023control} to directly manipulate the spatial representation of NeRF.
However, these methods either heavily rely on user interactions or have limitations in terms of spatial deformation and color transfer capabilities.

\begin{figure*}[t]
  \centering
   \includegraphics[width=1.0\linewidth]{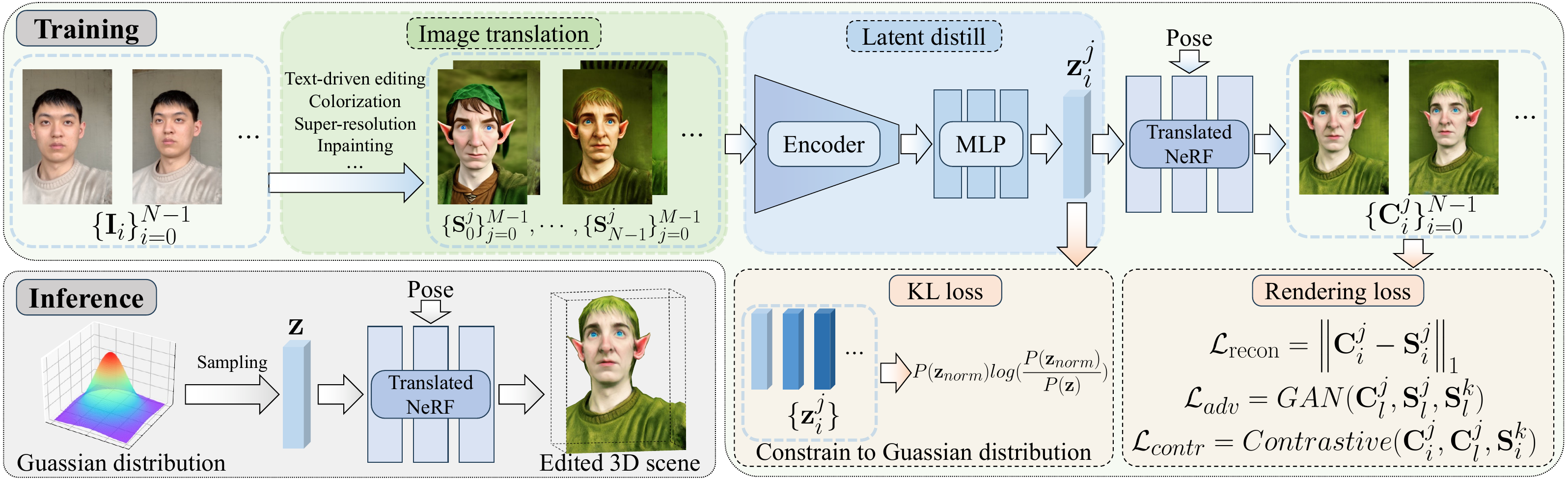}
   \caption{\textbf{Overview of GenN2N.} We first edit the source image set $\{ \mathbf{I}_i\}^{N-1}_{i=0}$ using 2D image-to-image translation methods, e.g., text-driven editing, colorization, zoom out, etc. For each view $i\in [0,N-1]$, we generate $M$ edited images, resulting in a group of translated image set $\{ \{ \mathbf{S}^j_i\}^{M-1}_{j=0} \}^{N-1}_{i=0}$. Then we use the Latent Distill Module to learn $M \times N$ edit code vectors from the translated image set, which serve as the input of the translated NeRF. To optimize our GenN2N, we design four loss functions: a KL loss to constrain the latent vectors to a Gaussian distribution; and $\mathcal{L}_{\textrm{recon}}$, $\mathcal{L}_{\textrm{adv}}$ and $\mathcal{L}_{\textrm{contr}}$ to optimize the appearance and geometry of the translated NeRF. At inference, we can sample a latent vector $\mathbf{z}$ from Gaussian distribution and render a corresponding multi-view consistent 3D scene with high quality.}
   \label{fig:pipeline}
\end{figure*}

\noindent{\bf NeRF Stylization.} 
% In the context of NeRF stylization, NeRFEditor~\citep{sun2022NeRFeditor} and StylizedNeRF~\citep{huang2022stylizedNeRF} employ mutual learning to maintain the 3D-consistent ability of NeRF when transferring 2D edits to 3D models. 
Images-referenced stylization~\citep{huang2022stylizedNeRF,chiang2022stylizing,zhang2022arf} often prioritize capturing texture style rather than detailed content, resulting in imprecise editing appearance of NeRF only. Text-guided works~\citep{wang2023nerf,wang2022clip}, on the other hand, apply contrastive losses based on CLIP~\citep{radford2021learning} to achieve the desired edits. While text references usually describe the global characteristics of the edited results, instructions offer a more convenient and precise expression. 
% Our method, compared with based on purely language-based instructions, supports controllable editing of NeRF's shape and appearance without additional operations.

\noindent{\bf Instruct-driven NeRF editing.} Among numerous image-to-image translation works, InstructPix2Pix~\citep{brooks2022instructpix2pix} stands out by efficiently editing images following instructions. It leverages large pre-trained models in the language and image domains~\citep{brown2020language,rombach2022high} to generate paired data (before and after editing) for training. While editing NeRF solely based on edited images is problematic due to multi-view inconsistency. To address this, an intuitive yet heavy approach~\citep{haque2023instruct} is to iteratively edit the image and optimize NeRF. In addition, NeRF-Art~\citep{wang2023nerf} and DreamEditor~\citep{zhuang2023dreameditor} adopt a CLIP-based contrastive loss~\citep{radford2021learning} and score distillation sampling~\citep{poole2022dreamfusion} separately to supervise the optimization of editing NeRF. Inspired by Generative Radiance Fields~\citep{schwarz2020graf,chan2022efficient}, We capture various possible NeRF editing in the generative space to solve it.

\section{Method}
\label{headings}

Given a NeRF scene, we present a unified framework GenN2N to achieve various editing on the 3D scene leveraging geometry and texture priors from 2D image editing methods, such as text-driven editing, colorization, super-resolution, inpainting, etc. 
While a universal image-to-image translator can theoretically accomplish these 2D editing tasks, we actually use a state-of-the-art translator for each task. 
Therefore, we formulate each 2D image editing method as a plug-and-play image-to-image translator and all NeRF editing tasks as our universal NeRF-to-NeRF translation, in which the given NeRF is translated into NeRF scenes with high rendering quality and 3D geometry consistency according to the user-selected editing target. 
The overview of GenN2N is illustrated in Fig.~\ref{fig:pipeline}, we first perform image-to-image translation in the 2D domain and then lift 2D edits to 3D and achieve NeRF-to-NeRF translation. 

Given $N$ multi-view images $\{\mathbf{I}_i\}_{i=0}^{N-1}$ of a scene, we first use Nerfstudio~\citep{nerfstudio} to train the original NeRF. Then we use a plug-and-play image-to-image translator to edit these source images. However, the content generated by the 2D translator may be inconsistent among multi-view images. For example, using different initial noise, the 2D translator~\citep{avrahami2023blended} may generate different content for image editing, which makes it difficult to ensure the 3D consistency between different view directions in the 3D scene. 
% To ensure the 3D consistency and rendering quality, we propose to model the distribution of the underlying 3D edits through a generative model that can cover all possible edited NeRFs, by learning an edit code for each edited image so that the generated content can be controlled by this edit code during the NeRF-to-NeRF translation process. 

To ensure the 3D consistency and rendering quality, we propose to model the distribution of the underlying 3D edits through a generative model that can cover all possible edited NeRFs, by learning an edit code for each edited image so that the generated content can be controlled by this edit code during the NeRF-to-NeRF translation process. 

For each view $\in [0, N-1]$, we generate $M$ edited images, resulting in a group of the translated image set $\{\{ \mathbf{S}^j_i\}^{M-1}_{j=0} \}^{N-1}_{i=0}$. Then we design a Latent Distill Module described in Sec.~\ref{method: latent distill module} to map each translated image $\mathbf{S}^j_i$ into an edit code vector $\mathbf{z}^j_i$ and design a KL loss $\mathcal{L}_{\textrm{KL}}$ to constrain those edit code vectors to a Gaussian distribution. Conditioned on the edit code $\mathbf{z}^j_i$, we perform NeRF-to-NeRF translation in Sec.~\ref{method: NeRF-to-NeRF translation} by rendering multi-view images $\{\mathbf{C}_i\}^{N-1}_{i=0}$ and optimize the translated NeRF by three loss functions: the reconstruction loss $\mathcal{L}_{\textrm{recon}}$, the adversarial loss $\mathcal{L}_{\textrm{AD}}$, and the contrastive loss $\mathcal{L}_{\textrm{contr}}$. After the optimization of the translated NeRF, as described in Sec.~\ref{method: inference}, we can sample an edit code $\mathbf{z}$ from Gaussian distribution and render the corresponding edited 3D scene with high quality and multi-view consistency in the inference stage.

\begin{figure}[t]
  \centering
   \includegraphics[width=0.9\linewidth]{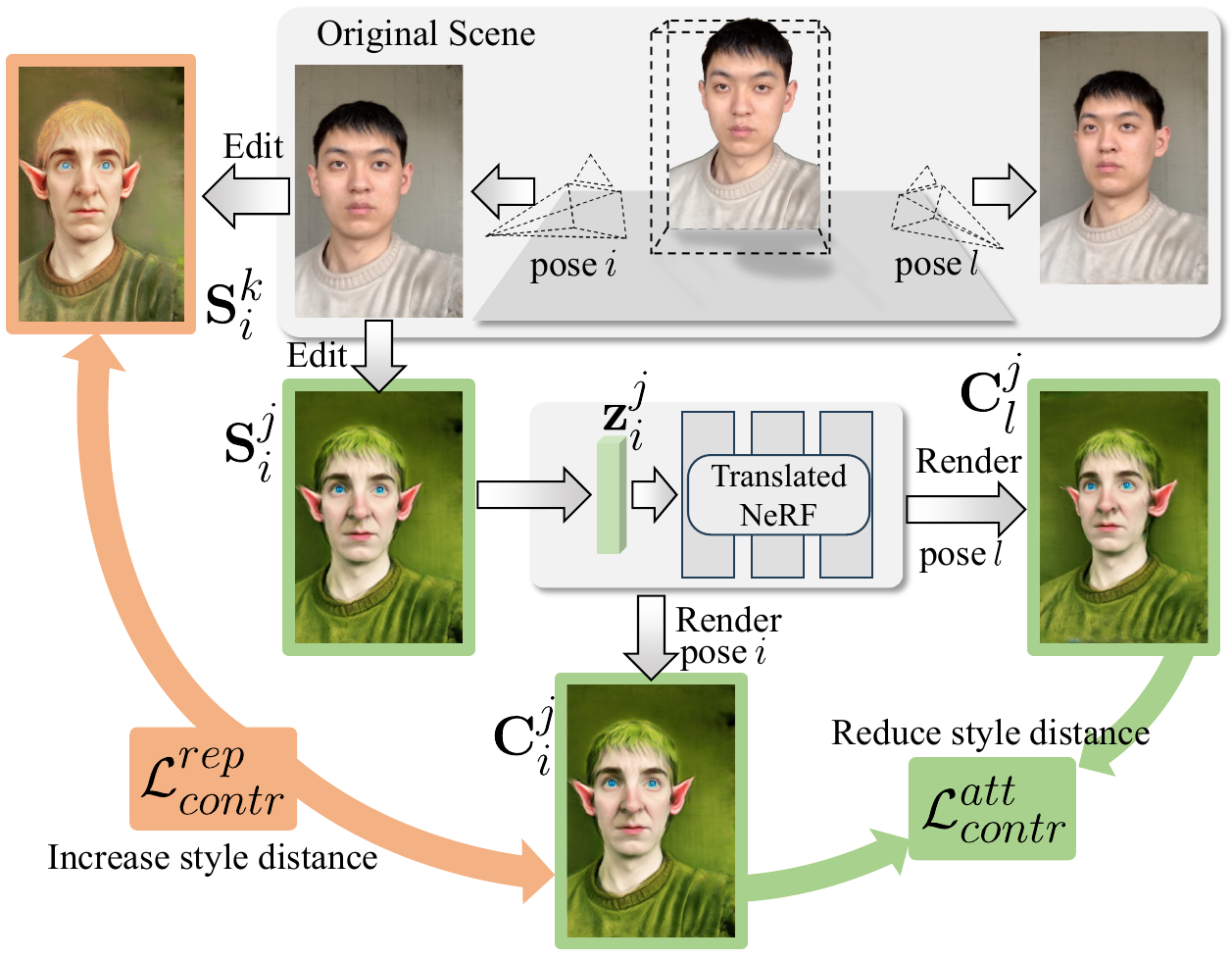}
   \caption{\textbf{Illustration of our proposed contrastive loss functions.} Regarding the multi-view rendered images $\mathbf{C}_i^j$ and $\mathbf{C}_l^j$ sharing the same edit code, we resend them to our Latent Distill Module to extract ${\mathbf{z}}_i^j$ and ${\mathbf{z}}_l^j$, and aggregate them via $\mathcal{L}_{\textrm{contr}}^{\textrm{att}}$. In addition, for $\mathbf{S}_i^k$ whose editing style vary from $\mathbf{S}_i^j$, $\mathcal{L}_{\textrm{contr}}^{\textrm{rep}}$ increase the distance between edit codes of them.}
   \label{fig:contrastive loss}
\end{figure}

\begin{figure}[t]
  \centering
  %\fbox{\rule{0pt}{3in} \rule{0.9\linewidth}{0pt}}
   \includegraphics[width=0.95\linewidth]{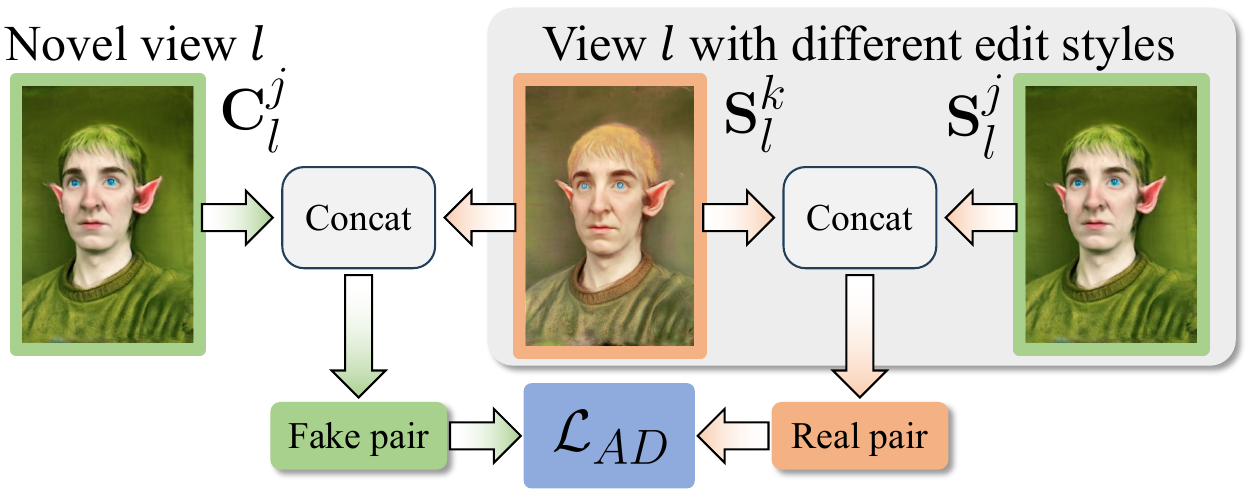}
   \caption{\textbf{Illustration of our proposed conditional adversarial loss functions.} Our conditional discriminator distinguishes artifacts such as blur and distortion in novel-view rendered image $\mathbf{C}_l^j$ compared with target image $\mathbf{S}_l^j$. $\mathbf{S}_l^j$ and $\mathbf{S}_l^k$ are edited with same view but various styles, the latter serves as the condition to concatenate with $\mathbf{C}_l^j$ and $\mathbf{S}_l^j$ and manufacture fake and real pairs.}
   \label{fig:gan loss}
\end{figure}

\subsection{Latent Distill Module}\label{method: latent distill module}

\noindent{\bf Image Translation.}
As illustrated in Fig.~\ref{fig:pipeline}, GenN2N is a unified framework for NeRF-to-NeRF translation, in which the core is to perform a 2D image-to-image translation and lift 2D edits into universal 3D NeRF-to-NeRF translation. 
Given the source multi-view image set $\{\mathbf{I}_i\}^{N-1}_{i=0}$ of a NeRF scene, we first perform image editing M times for each view using a plug-and-play 2D image-to-image translator, producing a group of translated image set $\{\{\mathbf{S}^j_i\}^{M-1}_{j=0} \}^{N-1}_{i=0}$. In this paper, we use several 2D translation tasks to show the adaptability of our GenN2N: text-driven editing,  super-resolution, colorization and inpainting. For more details about those 2D image editing methods, please refer to the supplementary materials. 

 % Here, we denote the 2D image-to-image translator as a generative model that can solve arbitrary 2D image editing tasks. 

\noindent{\bf Edit Code.}
Since 2D image-to-image translation may generate different content even with the same editing target, causing the inconsistency problem in the 3D scene. We propose to map each edited image $\mathbf{S}^j_i$ into a latent feature vector named edit code to characterize these diverse editings. We employ the off-the-shelf VAE encoder from stable diffusion~\citep{rombach2022high} to extract the feature from $\mathbf{S}^j_i$ and then apply a tiny MLP network to produce this edit code $\mathbf{z}^j_i \in \mathbb{R}^{64}$. During the training process, we keep the pre-trained encoder fixed and only optimize the parameters of the tiny MLP network. This mapping process can be formulated as follows:
\begin{equation}
\label{eqn:edit code mapping}
\mathbf{z}_i^j = \mathcal{D}(\mathbf{S}_i^j) = \mathcal{M}(\mathcal{E}(\mathbf{S}_i^j)),
\end{equation}
where $\mathcal{D}$ represent this mapping process, $\mathcal{E}$ is the fixed encoder, and $\mathcal{M}$ is the learnable tiny MLP.

\noindent{\bf KL loss.}
In order to facilitate effective sampling of the edit code so as to control the editing diversity of our NeRF-to-NeRF translation, we need to constrain the edit code to a well-defined distribution. Thus we design a KL loss to encourage $\mathbf{z}_i^j$ to approximate a Gaussian distribution:
\begin{equation}
\label{eqn:KL loss}
\mathcal{L}_{\textrm{KL}} = \mathbb{E}_{\mathbf{S} \in \{\{ \mathbf{S}^j_i\}^{M-1}_{j=0} \}^{N-1}_{i=0}} [P(\mathbf{z}_{normal})log(\frac{P(\mathbf{z}_{normal})}{P(\mathcal{D}(\mathbf{S}))})],
\end{equation}
where $\mathit{P({\mathbf{z}_{normal}})}$ denotes probability distribution of the standard Gaussian distribution in $\mathbb{R}^{64}$ and $\mathit{P(\mathcal{D}({\mathbf{S}}))}$ means probability distribution of the extracted edit codes.

\begin{figure*}[t]
  \centering
   \includegraphics[width=1.0\linewidth]{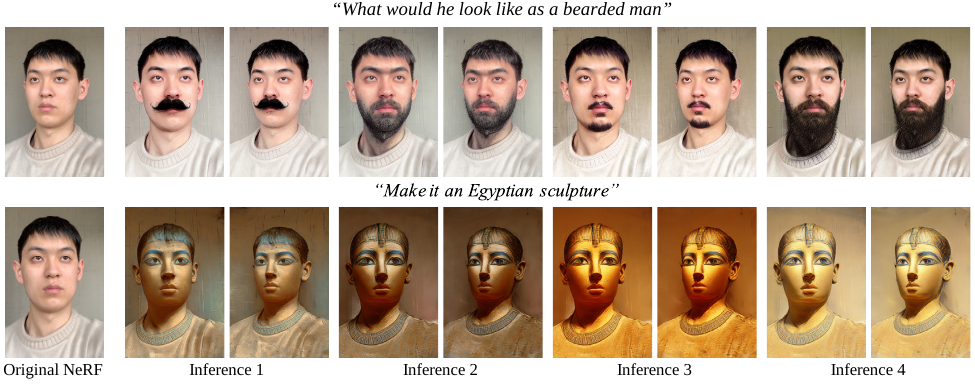}
   \caption{{\bf Text-Driven Editing.} We sample 4 inference results for both text-driven editing tasks. The diversity of geometry and appearance showcases awesome generative ability of GenN2N, on the premise of maintaining the 3D consistency between different viewpoints.}
   \label{fig:Editing}
\end{figure*}

\begin{figure*}[t]
  \centering
   \includegraphics[width=1.0\linewidth]{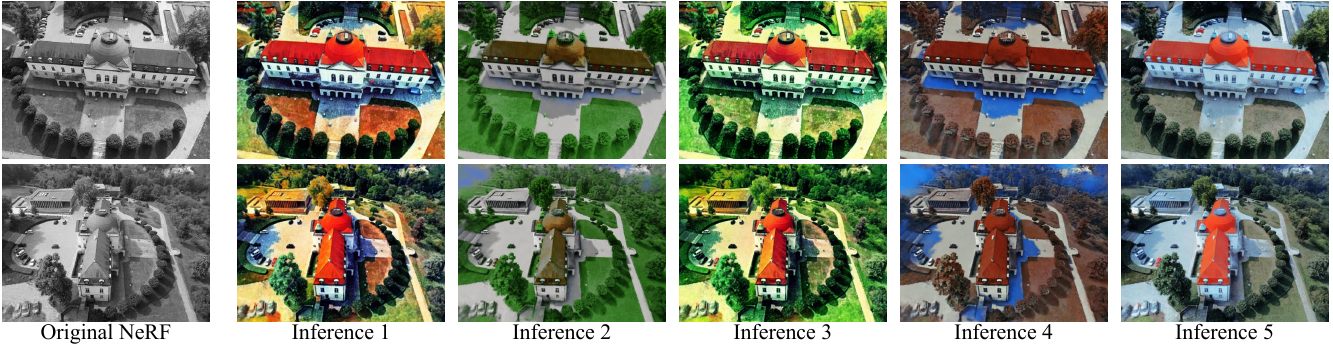}
   \caption{{\bf Colorization.} Our method colorizes the gray-scale 3D scene consistently across views. By changing the edit code during inference, diverse colorized scenes can be rendered with satisfying photorealism and reasonably rich colors.}
   \label{fig:Colorization}
\end{figure*}

\noindent{\bf Contrastive loss.}
It is not assured that edit codes $\mathbf{z}$ obtained from the Latent Distill Module contain only the editing information while excluding viewpoint-related effects. However, since the translated NeRF utilizes $\mathbf{z}$ to edit scenes, it yields instability if $\mathbf{z}$ violently changes given images that are similar in appearance but different in viewpoints. To ensure the latent code does not depend on 2D viewpoints but truly reflects the 3D edits, we regularize the latent code through a contrastive learning scheme. Specifically, we reduce the distance between edit codes of different-view rendered images from a translated NeRF that share the same edit code, while increasing the distance between same-view images that are multi-time edited by the 2D image-to-image translator. As illustrated in Fig.~\ref{fig:contrastive loss}, given an edit code $\mathbf{z}_i^j$ extracted from the $i$-th input view at the $j$-th edited image $\mathbf{S}_i^j$, we render multi-view images $\{ \mathbf{C}_l^j  \}_{l=0}^{N-1}$ using the translated NeRF conditioned on $\mathbf{z}_i^j$. Then we employ contrastive learning to encourage the edit code ${\mathbf{z}}_i^j$ to be close to $\{\mathbf{z}_l^j \}_{l=0}^{N-1}$ extracted from $ \{ \mathbf{C}_l^j \}_{l=0}^{N-1}$, while being distinct from the edit codes $\{\mathbf{z}_i^k \}_{k=0}^{M-1}$ extracted from $\{ \mathbf{S}_i^k  \}_{k=0}^{M-1}$, where $k\neq j$.

Specifically, our contrastive loss is designed as follows:
\begin{equation}
\begin{aligned}
\mathcal{L}_{\textrm{contr}} &= \mathcal{L}_{\textrm{contr}}^{\textrm{att}} + \mathcal{L}_{\textrm{contr}}^{\textrm{rep}} \\ 
&= \sum_{l=0}^{N-1}||{\mathbf{z}}^j_{i}-{\mathbf{z}}^j_{l}||_2^2 + \sum_{k=0}^{M-1}max(0,\alpha-||{\mathbf{z}}^j_{i}-{\mathbf{z}}_{i}^k||_2^2),
\end{aligned}
\end{equation}
where $\alpha$ represents the margin that encourages the difference in features, and $k\neq j$.

\subsection{NeRF-to-NeRF translation}\label{method: NeRF-to-NeRF translation}

\noindent{\bf Translated NeRF.}
After 2D image-to-image translation, we need to lift these 2D edits to the 3D NeRF. For this purpose, we propose to modify the original NeRF as a translated NeRF that takes the edit code $\mathbf{z}$ as input and generates the translated 3D scene according to the edit code. We refer readers to the supplementary for more details about the network architecture.

\noindent{\bf Reconstruction loss.}
Given an edit code $\mathbf{z}_i^j$ extracted from the edited image $\mathbf{S}_i^j$, we can generate a translated NeRF to render $\mathbf{C}_i^j$ from the same viewpoint. Then we define the reconstruction loss as the L1 normalization and Learned Perceptual Image Patch Similarity (LPIPS)~\citep{zhang2018unreasonable} between the rendered image $\mathbf{C}_i^j$ and the edited image $\mathbf{S}_i^j$ as follows:
 \begin{equation}
\begin{aligned}
\mathcal{L}_{\textrm{recon}} &= \mathcal{L}_{\textrm{L1}} + \mathcal{L}_{\textrm{LPIPS}} \\ 
&= \left \| \mathbf{C}_i^j - \mathbf{S}_i^j\right \|_{1} + LPIPS[\mathcal{P}(\mathbf{C}_i^j) - \mathcal{P}(\mathbf{S}_i^j)],
\end{aligned}
\end{equation}
where $\mathcal{P}$ means a patch sampled from the image. Note that due to the lack of 3D consistency of the edited multi-view image, the supervision of the edited image from other viewpoints $\{\mathbf{S}_l^j\}_{l \neq i}$ will lead to conflicts in pixel-space optimization. Therefore, we only employ reconstruction loss on the same view image $\mathbf{S}_i^j$ to optimize the translated NeRF.

\begin{figure}[t]
  \centering
   \includegraphics[width=1.0\linewidth]{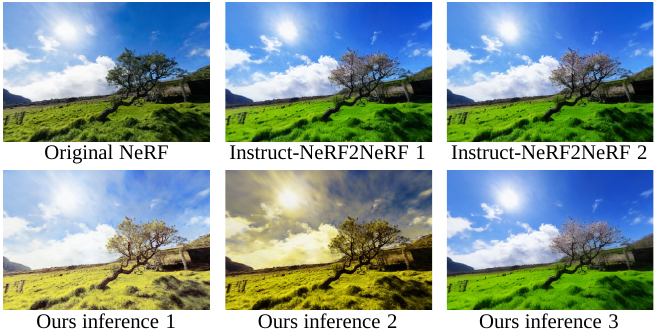}
   \caption{{\bf Comparisons with baselines of text-driven NeRF editing.} We compare our method with Instruct-NeRF2NeRF~\citep{haque2023instruct} in the editing by using the text prompt \textit{``Make it Spring''}.
   }
   \label{fig:Compare_text_editing}
\end{figure}

\begin{table}[t]
\centering
\resizebox{0.5 \textwidth}{!}{
    \begin{tabular}{
            >{\centering\arraybackslash}p{3.5cm} % Method 
            >{\centering\arraybackslash}p{3.0cm} % text-driven editing
            >{\centering\arraybackslash}p{2.5cm} % text-driven editing
            >{\centering\arraybackslash}p{1.0cm} % text-driven editing
        }%{|p{3cm}|p{3cm}|p{3cm}||p{3cm}|p{3cm}||p{3cm}|p{3cm}||p{3cm}|p{3cm}||p{3cm}|p{3cm}|}
        %\resizebox{\textwidth}{12mm}
        \toprule
        \multirow{2}{*}{Method} &CLIP Text-Image  & CLIP Direction & \multirow{2}{*}{FID $\downarrow$}  \\
        &Direction Similarity$\uparrow$  &Consistency $\uparrow$\\
        \midrule
       InstructPix2Pix~\citep{brooks2022instructpix2pix}+NeRF     & 0.1669  & 0.8475  & 270.542   \\
       % DDColor  & --  & --  & --   \\
       % ResShift & --  & --  & --   \\
       % BLD      & --  & --  & --   \\
        \midrule
       Instruct-NeRF2NeRF        & 0.2021  & 0.9828  & 148.021   \\
       % IN2N+DDColor & --  & --  & --   \\
       % IN2N+ResShift& --  & --  & --    \\
       % IN2N+BLD     & --  & --  & --   \\
        % \midrule
       % DreamEditor & aa  & aa  & aa  \\
       % PaletteNeRF & aa  & aa  & aa   \\
    %     \midrule
       % NeRF-SR     & --  & --  & --  \\
       % 4k-NeRF     & --  & --  & -- \\
    %     \midrule
       % SPin-NeRF   & --  & --  & --   \\
       % ROFNRF      & --  & --  & --  a  \\
        \midrule
        % \midrule
       Ours w/o $\mathcal{L}_{\textrm{adv}}$ & 0.1920  & 0.9657  & 162.275  \\
        % \midrule
       Ours w/o $\mathcal{L}_{\textrm{contr}}$ & 0.2007  & 0.9749  & 156.524   \\
       Ours  & \textbf{0.2089}  & \textbf{0.9864}  & \textbf{137.740}   \\
        \bottomrule
    \end{tabular}
}
    \caption{\textbf{Quantitative results on text-driven editing.} 
    % We choose InstructPix2Pix as 2D translator and conduct NeRF translation via traditional NeRF optimization, Instruct-NeRF2NeRF and our method.
    We compare our method with the naive method of directly combining InstractPix2Pix~\citep{brooks2022instructpix2pix} with NeRF and the state-of-the-art method Instruct-NeRF2NeRF~\citep{haque2023instruct}.  
    }
\label{tabel: comparision_table_editing}
\end{table}

\noindent{\bf Adversarial loss.}
Since the 3D consistency of edited multi-view images is not assured, relying solely on the reconstruction loss on the same view often leads to blurry or distorted artifacts on novel views.
Previous research demonstrates the effectiveness of conditional adversarial training in preventing the production of blurry rendered images resulting from conflicts that arise from noise in the camera extrinsic when performing image supervision from different viewpoints~\citep{huang2020adversarial}. The function of the condition is to guide discriminator with fine-grained information from the same viewpoint, thus preventing GAN mode collapse. 

It inspires us to incorporate conditional adversarial loss on rendered images from the translated NeRF, which is conducive to distinguish artifacts in rendered images. As illustrated in Fig.\ref{fig:gan loss}, the discriminator $\mathbf{D}$ takes into real pairs and fake pairs. Each real pair $\mathbf{R}$ consists of ${\mathbf{S}^j}$ and ${\mathbf{S}^j-\mathbf{S}^k}$ where ${\mathbf{S}^j \in  \{ \mathbf{S}_{i}^j  \}_{i=0}^{N-1}}$ and ${\mathbf{S}^k \in  \{ \mathbf{S}^k_{i}  \}_{i=0}^{N-1}}$ are from two sets of edited images from the image translation. Similarly, each fake pair $\mathbf{F}$ consists of ${\mathbf{C}^j}$ and ${\mathbf{C}^j-\mathbf{S}^k}$ in which ${\mathbf{C}^j \in  \{ \mathbf{C}_{i}^j  \}_{i=0}^{N-1}}$ is generated by translated NeRF. Note that the images in the same pair come from the same viewpoint. The pairs are concatenated in RGB channels and fed into the discriminator. We optimize the discriminator $\mathbf{D}$ and translated NeRF with the objective functions below:
\begin{equation}
\begin{aligned}
 \mathcal{L}_{\textrm{AD-D}} &= \mathbb{E}_{\mathbf{R}}[-log(\mathbf{D}(\mathbf{R}))]+\mathbb{E}_{\mathbf{F}}[-log(1-\mathbf{D}(\mathbf{F}))], \\ 
\mathcal{L}_{\textrm{AD-G}} &= \mathbb{E}_{\mathbf{F}}[-log(\mathbf{D}(\mathbf{F}))]. \\
\end{aligned}
\end{equation}

\begin{figure}[t]
  \centering
   \includegraphics[width=1.0\linewidth]{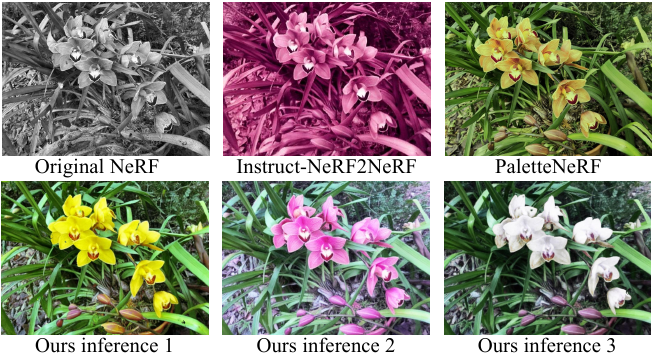}
   \caption{{\bf Comparisons with baselines of NeRF colorization.} We compare with PaletteNeRF\cite{kuang2023palettenerf} in colorization.
   }
   \label{fig:Compare_color}
\end{figure}

\begin{table}[t]
\centering
\resizebox{0.5 \textwidth}{!}{
    \begin{tabular}{
            >{\centering\arraybackslash}p{3.5cm} % Method 
            % >{\centering\arraybackslash}p{3.0cm} % text-driven editing
            >{\centering\arraybackslash}p{2.5cm} % text-driven editing
            >{\centering\arraybackslash}p{2.5cm} % text-driven editing
        }%{|p{3cm}|p{3cm}|p{3cm}||p{3cm}|p{3cm}||p{3cm}|p{3cm}||p{3cm}|p{3cm}||p{3cm}|p{3cm}|}
        %\resizebox{\textwidth}{12mm}
        \toprule
        Method &CF $\uparrow$ & FID $\downarrow$  \\
        \midrule
       DDColor~\citep{kang2022ddcolor}+NeRF     & 40.435  & 148.957   \\
       % DDColor  & --  & --  & --   \\
       % ResShift & --  & --  & --   \\
       % BLD      & --  & --  & --   \\
        \midrule
       % IN2N         & aa  & aa  & aa   \\
       Instruct-NeRF2NeRF  & 45.599  & 201.456   \\
       % IN2N+ResShift& --  & --  & --    \\
       % IN2N+BLD     & --  & --  & --   \\
    %     \midrule
       % DreamEditor & aa  & aa  & aa  \\
       PaletteNeRF~\citep{kuang2023palettenerf} & 39.654  & --    \\
    %     \midrule
       % NeRF-SR     & --  & --  & --  \\
       % 4k-NeRF     & --  & --  & -- \\
    %     \midrule
       % SPin-NeRF   & --  & --  & --   \\
       % ROFNRF      & --  & --  & --  a  \\
        \midrule
        % \midrule
       Ours w/o $\mathcal{L}_{\textrm{adv}}$ & 35.031  & 137.740   \\
        % \midrule
       Ours w/o $\mathcal{L}_{\textrm{contr}}$ & 34.829  & 105.750    \\
       Ours & \textbf{65.099}  &  \textbf{35.041}   \\
        \bottomrule
    \end{tabular}
}
    \caption{\textbf{Quantitative results on colorization.} We colorize images with the translator, DDcolor~\citep{kang2022ddcolor}, and edit NeRF by directly optimizing NeRF, Instruct-NeRF2NeRF~\citep{haque2023instruct} and our NeRF translation method. The quantitative comparison is conducted between these methods as well as PaletteNeRF~\citep{kuang2023palettenerf}.}
\label{tabel: comparision_table_color}
\end{table}

\noindent{\bf Optimization.}
During the training process, we jointly optimize the loss functions mentioned above: $\mathcal{L}_{\textrm{KL}}$ and $\mathcal{L}_{\textrm{contr}}$ for the edit code, $\mathcal{L}_{\textrm{recon}}$ and $\mathcal{L}_{\textrm{AD-G}}$ for the translated NeRF, and $\mathcal{L}_{\textrm{AD-D}}$ for the discriminator. The total loss formula is expressed as follows: 
\begin{equation}
\mathcal{L} = \mathcal{L}_{\textrm{KL}} + \mathcal{L}_{\textrm{recon}}  + 
\mathcal{L}_{\textrm{AD-G}} + 
\mathcal{L}_{\textrm{AD-D}} + 
\mathcal{L}_{\textrm{contr}}.
\end{equation}
where we assign each regularization term the weight of 1.0, 1.0, 0.1, 0.1, 0.1 in all of our experiments. 
The weights can be adjusted to prioritize different aspects of the training objective, such as reconstruction accuracy, adversarial training, and perceptual quality.

\subsection{Inference}\label{method: inference}
After the optimization of our GenN2N, the translated NeRF is optimized to be able to render the target scene conditioned on the edit code. As shown in Fig.~\ref{fig:teaser}, users can simply sample an edit code from the Gaussian distribution and use the translated NeRF to render the 3D scene with high-quality and multi-view 3D consistency.

\section{Experiments}
\label{sec:experiments}

% \begin{figure}[t]
%   \centering
%    \includegraphics[width=0.8\linewidth]{Figures/exp/inpainting-tmp.png}
%    \caption{{\bf Inpainting.} }
%    \label{fig:Inapinting}
% \end{figure}

%We compare with SPIn-NeRF~\citep{mirzaei2023spin}in object removal on the data provided by SPIn-NeRF. Our method preserves details of the scene while successfully eliminating the object, while SPInNeRF~\citep{mirzaei2023spin} causes the surrounding scene to blur and deform.  

%**************************************************************
Our proposed GenN2N is a unified NeRF-to-NeRF translation framework which can support various NeRF editing tasks. In this paper, we demonstrate the effectiveness of GenN2N by a suite of challenging tasks: %NeRF-to-NeRF translation
% \vspace{-.15cm}
\begin{enumerate}[label={(\arabic*)}, topsep=0.1pt, partopsep=0pt, leftmargin=20pt, parsep=0pt, itemsep=0.1pt]
    \item \textbf{Text-driven Editing} edits the given NeRF scene to a set of NeRF scenes according to the text instruction.
    \item \textbf{Colorization} transforms a gray-scale NeRF scene to a set of plausible color NeRF scenes.
    \item \textbf{Super-resolution} enhances the resolution of NeRF and enables multiple plausible outcomes. 
    % \item \textbf{Object Removal} removes the target object in the NeRF scene while keeping other contents, especially the background content, unchanged and plausible.
    % \item \textbf{Zoom Out} extends an input NeRF along the input region to enlarge NeRF scenes.
    \item \textbf{Inpainting} fills in user-specified masked regions in the NeRF scene with realistic content.
\end{enumerate}
We achieve those tasks by simply changing the plug-and-play 2D image translator in our framework, without any additional task-specific design.
Previous studies have extensively explored some of these issues like text-driven editing, colorization, super-resolution, and inpainting. However, there is rarely a unified framework that can achieve all these tasks with strong performance, high quality, and plausible multi-view consistent 3D structure.
Furthermore, GenN2N can also perform zooming out and text-driven inpainting in NeRF-to-NeRF translation, which were not explored in prior research. We refer readers to the supplementary materials for detailed experiment settings, dataset settings and implementation details.

\begin{figure}[t]
  \centering
   \includegraphics[width=1.0\linewidth]{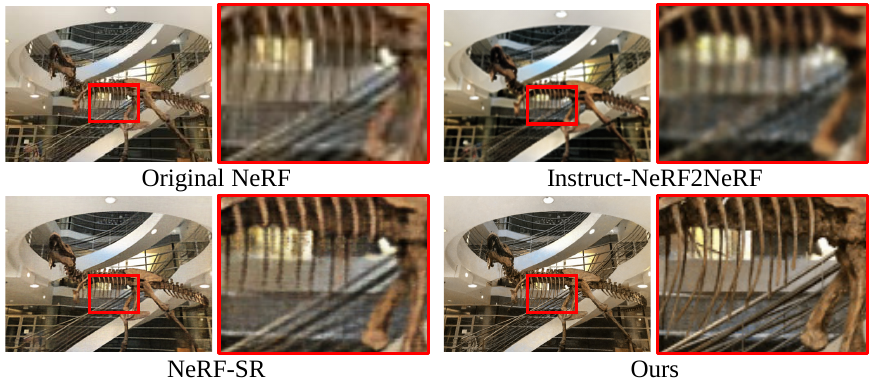}
   \caption{{\bf Comparisons with baselines of NeRF super-resolution.} We compare with NeRF-SR~\citep{wang2022nerf} in super-resolution.
   }
   \label{fig:Compare_sr}
\end{figure}

\begin{table}[t]
\centering
\resizebox{0.5 \textwidth}{!}{
    \begin{tabular}{
            >{\centering\arraybackslash}p{3.5cm} % Method 
            >{\centering\arraybackslash}p{2.5cm} % text-driven editing
            >{\centering\arraybackslash}p{2.5cm} % text-driven editing
            >{\centering\arraybackslash}p{2.5cm} % text-driven editing
        }%{|p{3cm}|p{3cm}|p{3cm}||p{3cm}|p{3cm}||p{3cm}|p{3cm}||p{3cm}|p{3cm}||p{3cm}|p{3cm}|}
        %\resizebox{\textwidth}{12mm}
        \toprule
        Method &PSNR $\uparrow$ &SSIM $\uparrow$ &LPIPS $\downarrow$ \\
        \midrule
       % DDColor     & aa  & aa   \\
       % DDColor  & --  & --  & --   \\
       ResShift~\citep{yue2023resshift}+NeRF & 19.978  & 0.535  & 0.1156   \\
       % BLD      & --  & --  & --   \\
        \midrule
       % IN2N         & aa  & aa  & aa   \\
       % IN2N+DDColor & aa  & aa   \\
       Instruct-NeRF2NeRF & 20.299  & 0.642  & 0.2732    \\
       % IN2N+BLD     & --  & --  & --   \\
    %     \midrule
       % DreamEditor & aa  & aa  & aa  \\
       % PaletteNeRF & aa  & aa    \\
    %     \midrule
       NeRF-SR~\citep{wang2022nerf}     & 27.957  & 0.897  & 0.0937  \\
       % 4k-NeRF     & --  & --  & -- \\
    %     \midrule
       % SPin-NeRF   & --  & --  & --   \\
       % ROFNRF      & --  & --  & --  a  \\
        \midrule
        % \midrule
       Ours w/o $\mathcal{L}_{\textrm{adv}}$ & 12.555  & 0.663 & 0.2001   \\
        % \midrule
       Ours w/o $\mathcal{L}_{\textrm{contr}}$ & 15.372  & 0.662  & 0.1834   \\
       Ours  & \textbf{28.501}  & \textbf{0.913} &\textbf{0.0748}   \\
        \bottomrule
    \end{tabular}
}
    \caption{\textbf{Quantitative results on super-resolution.} We improve image resolution with ResShift~\citep{yue2023resshift} and edit NeRF by directly optimizing NeRF, Instruct-NeRF2NeRF~\citep{haque2023instruct} and our NeRF translation method. The quantitative comparison is conducted between these methods as well as NeRF-SR~\citep{wang2022nerf}.}
\label{tabel: comparision_table_SR}
\end{table}

\subsection{Comparisons}

\noindent{\bf Text-driven Editing.}
We achieve text-driven editing of the given NeRF by using InstructPix2Pix~\citep{brooks2022instructpix2pix} as the 2D image-to-image translator in our framework. We compared our approach to a naive solution, which involves optimizing a NeRF with edited images via InstructPix2Pix. However, this naive approach leads to a 3D inconsistency problem among different edits. While Instruct-NeRF2NeRF~\citep{haque2023instruct} proposed an iterative updating mechanism to address this issue, it falls short in capturing the diversity of different edits, making it challenging to ensure the quality of the outcomes. To evaluate our method, we conducted experiments on the \textit{Face}~\citep{haque2023instruct} and \textit{Fangzhou}~\citep{wang2023nerf} self-portrait datasets, and \textit{Farm}~\citep{haque2023instruct} dataset, comparing GenN2N with the state-of-the-art NeRF editing method Instruct-NeRF2NeRF~\citep{haque2023instruct}. 

Quantitative results are presented in Table~\ref{tabel: comparision_table_editing}, where we employed CLIP Text-Image Direction Similarity~\citep{haque2023instruct}, CLIP Direction Consistency~\citep{haque2023instruct}, and Fréchet Inception Distance (FID)~\citep{heusel2017gans} as evaluation metrics. The results highlight the superior performance of GenN2N over other methods, demonstrating its effectiveness in producing high-quality 3D text-driven editing results.
Furthermore, we provide a qualitative comparison between GenN2N and Instruct-NeRF2NeRF~\citep{haque2023instruct} in Figure~\ref{fig:Compare_text_editing}. Notably, Instruct-NeRF2NeRF was trained twice, but the results are nearly identical. In contrast, GenN2N can render edited scenes with various effects, consistent with the input text instruction, by inferring with different edit codes sampled from a Gaussian distribution. To further illustrate the generative capability of GenN2N, we showcase additional results of our inference renderings in Figure~\ref{fig:Editing}, demonstrating its diverse generative ability in terms of appearance and geometry.

\noindent{\bf Colorization.}
For NeRF colorization, GenN2N uses DDColor~\citep{kang2022ddcolor} as the 2D image-to-image translator. CoRF~\cite{Dhiman2023CoRFC} and Palette-NeRF~\cite{kuang2023palettenerf} do a similar task and we compare with Palette-NeRF in Table~\ref{tabel: comparision_table_color}. The Fréchet Inception Distance (FID)~\citep{heusel2017gans} and colorfulness score (CF)~\cite{larsson2016learning} are used to measure the distribution similarity and vividness of generated images. 
We show visual comparison results in Fig.~\ref{fig:Colorization} and Fig.~\ref{fig:Compare_color}. We can find that with different edit codes, the scene can be rendered in different color styles. It is noticeable that with the same edit code, the color rendered from different views is consistent. This strongly demonstrates the effectiveness of our method in translating NeRF while keeping the 3D consistency of the scene. 

\noindent{\bf Super-resolution.}
When only low-resolution images are available, our methods can boost NeRF in reconstructing scenes at higher resolution, while keeping view consistency and avoiding blurry outputs. We achieve this by employing ResShift~\citep{yue2023resshift} as the 2D image-to-image translator in GenN2N. Following state-of-the-art method NeRF-SR~\citep{wang2022nerf}, we conduct experiments on \textit{LLFF} dataset~\citep{mildenhall2021nerf}, using PSNR, SSIM, and LPIPS as evaluation metrics. As shown in Table~\ref{tabel: comparision_table_SR}, GenN2N obtains NeRF-to-NeRF translation with higher performance than NeRF-SR~\cite{wang2022nerf}. Moreover, we also provide qualitative comparison results in Fig.~\ref{fig:Compare_sr}, where GenN2N produces clearer and more realistic rendering results than previous methods.

\begin{figure}[t]
  \centering
   \includegraphics[width=1.0\linewidth]{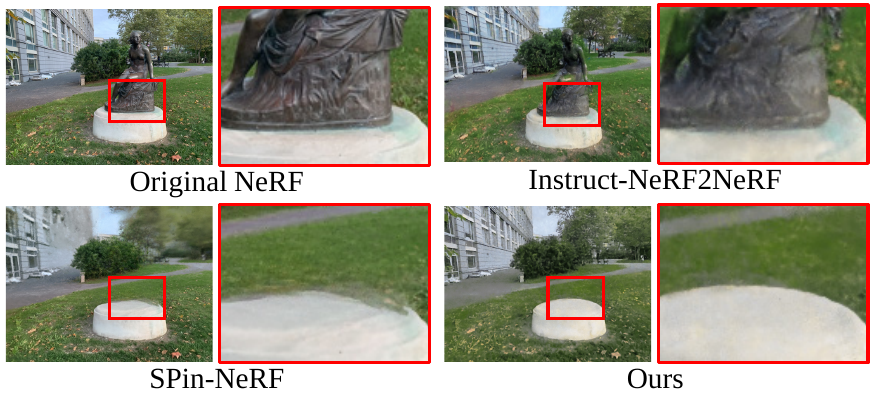}
   \caption{{\bf Comparisons with baselines of NeRF inpainting.} We compare with SPIn-NeRF~\citep{mirzaei2023spin} in inpainting. 
   }
   \label{fig:Compare_inpainting}
\end{figure}

\noindent{\bf Inpainting.}
The goal of NeRF Inpainting is to fill the 3D content of regions specified by users. SPIn-NeRF~\citep{mirzaei2023spin} achieves this through a multi-step process: it employs SAM~\citep{kirillov2023segment} for object segmentation, utilizes LaMa~\citep{suvorov2021resolution} to paint the background content in multi-view images, and subsequently trains the NeRF model with color, depth, and perceptual cues.
In our experiments, we use SAM and LaMa as the 2D image-to-image translator in our GenN2N, which is the same setting as SPin-NeRF~\cite{mirzaei2023spin}. Quantitative comparisons on \textit{statue} dataset~\citep{mirzaei2023spin} are shown in Table~\ref{tabel: comparision_table_inpainting}, where GenN2N achieves superior PSNR and SSIM scores than SPin-NeRF~\cite{mirzaei2023spin}, highlighting the effectiveness of our GenN2N framework. In addition, qualitative results are showcased in Fig.~\ref{fig:Compare_inpainting} revealing that while SPin-NeRF~\cite{mirzaei2023spin} fails to generate reasonable content behind the masked object, our GenN2N produces realistic content in the same region with fine multi-view consistency. 

\begin{table}[t]
\centering
\resizebox{0.5 \textwidth}{!}{
    \begin{tabular}{
            >{\centering\arraybackslash}p{3.5cm} % Method 
            >{\centering\arraybackslash}p{2.5cm} % text-driven editing
            >{\centering\arraybackslash}p{2.5cm} % text-driven editing
            >{\centering\arraybackslash}p{2.5cm} % text-driven editing
        }%{|p{3cm}|p{3cm}|p{3cm}||p{3cm}|p{3cm}||p{3cm}|p{3cm}||p{3cm}|p{3cm}||p{3cm}|p{3cm}|}
        %\resizebox{\textwidth}{12mm}
        \toprule
        Method &PSNR $\uparrow$ &SSIM $\uparrow$ &LPIPS $\downarrow$ \\
        \midrule
       LaMa~\citep{suvorov2021resolution}+NeRF     & 18.983  & 0.3706 & 0.1730   \\
        \midrule
       Instruct-NeRF2NeRF & 16.734  & 0.3088 &0.2750 \\
       SPin-NeRF~\citep{mirzaei2023spin}   & 24.369  & 0.7217 & 0.1754 \\
        \midrule
       Ours  & \textbf{26.868}  & \textbf{0.8137} & \textbf{0.1284}   \\
        \bottomrule
    \end{tabular}
}
    \caption{\textbf{Quantitative results on NeRF inpainting.}}
\label{tabel: comparision_table_inpainting}
\end{table}

\subsection{Ablation Studies}

We conduct comprehensive ablation experiments to validate the designs of each component in GenN2N. Due to space limitations, we only highlight the essential aspects below. Please refer to supplementary for more details.

\noindent{\bf The Contrastive Loss.} We demonstrate the advantages of incorporating our proposed contrastive loss in Table~\ref{tabel: comparision_table_editing}, \ref{tabel: comparision_table_color}, \ref{tabel: comparision_table_SR}. The motivation is to disentangle the camera view and edit information present in the latent space. 
We achieve this by reducing the distance between edit codes of different-view rendered images from a translated NeRF that shares the same edit code, while increasing the distance between same-view images that are edited by the 2D translator with diverse edit styles.
As demonstrated in Tables, the absence of contrastive loss leads to the generation of blurry areas in the rendered images, resulting in a decrease in the metric scores. This blurriness can be attributed to the inclusion of pose information within the edit code $\mathbf{z}$. By incorporating the contrastive loss, our method successfully achieves a uniform appearance with different observing views under the same style latent $\mathbf{z}$. 

\noindent{\bf Discriminator for Novel Views.} We demonstrate the effectiveness of employing a conditional discriminator to address artifacts caused by inconsistent cross-view edited images and to enhance the quality of novel view rendering images, as depicted in Table~\ref{tabel: comparision_table_editing}, \ref{tabel: comparision_table_color}, \ref{tabel: comparision_table_SR}. The removal of this conditional discriminator results in blurry novel view images with artifacts in the background region. We attribute these undesirable effects to the inability of current image-to-image translation methods, such as InstructPix2Pix~\cite{brooks2022instructpix2pix}, to produce image editing consistently across multi-view images. To mitigate these issues, we introduce a conditional discriminator between rendered images from the translated NeRF and edited images from the 2D image-to-image translator. This inclusion successfully eliminates artifacts and enhances the image quality of rendered images from the translated NeRF.

\subsection{Applications}

We demonstrate the versatility and robustness of GenN2N by exploring two translation applications: Zoom Out and Text-Driven Inpainting. While existing 2D translators~\citep{saharia2022palette, avrahami2023blended} can complete these tasks, 3D editing has not been explored. We achieve these tasks by incorporating Blended Latent Diffusion~\citep{avrahami2023blended} as the 2D image-to-image translator, enabling us to generate diverse and high-quality content with multi-view consistency. Please refer to our supplementary materials for the results of these applications due to space constraints.

\section{Conclusions}
We introduce GenN2N, a unified NeRF-to-NeRF translation framework that can handle various NeRF editing tasks. Unlike previous task-specific approaches, our framework uses an image-to-image translator for 2D editing and integrates the results into 3D NeRF space. To address the challenge of ensuring 3D consistency, we propose modeling the distribution of 3D edited NeRFs from 2D edited images using our novel techniques. After optimization, users can sample from the conditional generative model to obtain diverse 3D editing results with high rendering quality and multi-view consistency. Our experiments demonstrate that GenN2N outperforms existing task-specific methods on various editing tasks, including text-driven editing, colorization, super-resolution, and inpainting, in terms of efficiency, quality, and diversity.

% {
%     \small
%     \bibliographystyle{ieeenat_fullname}
%     \bibliography{main}
% }

% \begin{appendices}
% \end{appendices}
\newpage
\clearpage
\appendix
\twocolumn[{%
\centering
\title{\Large \textbf{Supplementary Material for GenN2N: Generative NeRF2NeRF Translation}}
\vspace{20pt}
}]

To make our \ourname{} self-contained, we provide more details in this document, including:

\begin{itemize}
\item More details about our method, including 2D image-to-image translator used in our pipeline and the architecture details of our translated NeRF.
\item Detailed settings of our experiments, including datasets settings and implementation details.
\item More insight experiments of our method, including quality verification of our generation space, comparisons with naive altering parameters in InstructNeRF2NeRF, interpolation of the edit code, and ablation study on hyperparameter $M$.
\item Additional experiment results, including more qualitative and quantitative results and additional applications of our \ourname.
\end{itemize}

\section{Method Details}

\subsection{2D image-to-image Translator}
In our proposed \ourname{}, we use plug-and-play image-to-image translators to perform editing on the 2D domain and optimize the translated NeRF to lift these 2D edits into the 3D NeRF space. Note that the 2D translator in our pipeline can be changed to support various of NeRF editing tasks, here for convenience of comparing our method with existing task-specific NeRF editing baselines, we different 2D translators to achieve corresponding editing tasks as follows:
\begin{itemize}
\item \noindent{\bf Text-driven Editing.} To achieve NeRF editing under text instructions, we use InstructPix2Pix~\citep{brooks2022instructpix2pix} as the 2D image-to-image translator in our framework. InstructPix2Pix is a diffusion-based method designed for image editing according to user-provided instructions. Specifically, InstructPix2Pix learns a U-Net to perform denoise diffusion to generate the target edited image based on the given image and the text embedding. While InstructPix2Pix can produce high-quality editing results that highly align with the input instructions, given different initial noise or input image, different content may be generated during the editing process of InstructPix2Pix, which makes it difficult to ensure the 3D consistency in the text-driven NeRF Editing process. 

\item \noindent{\bf Super-resolution.} For the NeRF super-resolution task, we choose ResShift~\citep{yue2023resshift} instead of InstructPix2Pix~\citep{brooks2022instructpix2pix} as the 2D image-to-image translator in GenN2N due to the unrobustness of InstructPix2Pix~\citep{brooks2022instructpix2pix} in the super-resoluion task. ResShift~\citep{yue2023resshift} is the current state-of-the-art image super-resolution method designed based on diffusion model. With dedicated designs for image super-resolution, such as the residual shifting mechanism and the flexible noise schedule, ResShift~\citep{yue2023resshift} can produce super-resolution images with high-quality. Thus, given a set of multi-view images of a NeRF scene, we directly use ResShift~\citep{yue2023resshift} to increase the resolution of all these images by the same factor of $\times 4$ as NeRF-SR~\citep{wang2022nerf}. 
% \kunming{Are those operations the same as previsous methods such as our baselines? Is the 2D editor also the same as our baseline, so that we can say that our method is more powerful than others just because of our design?}

\item \noindent{\bf Inpainting.} For the task of inpainting in NeRF, we aim to replace a certain region in a 3D scene, usually an object, and keep painted contents visually
plausible and consistent with the remained context. Following SPIn-NeRF~\citep{mirzaei2023spin}, we use LaMa~\citep{suvorov2022resolution} as our 2D image-to-image translator. The input of LaMa is an image and a binary mask that indicates the region to paint. 
% We use the segment anything model~\citep{kirillov2023segment}to generate the mask of the object at a specified location for multi-view images. 
We support various ways to get a mask, but note that multi-view masks must correspond to the same location in the 3D scene. For example, by artificially calculating the position of the part to paint in the 3D scene corresponding to the 2D image, or using the segment anything model~\citep{kirillov2023segment} to get the mask of the same object.
Based on these masked images, LaMa can successfully generate contents in the desired region that remain close to the input image with plausible 3D appearance and geometry. 
% \kunming{Are those operations the same as previsous methods such as our baselines? Yes, Following SPinNERF...}

\item \noindent{\bf Colorization.} To achieve 3D NeRF colorization, we use DDColor~\citep{kang2022ddcolor} as our plug-and-play 2D image-to-image translaton. Specifically, given a set of gray-scale multi-view images of a NeRF scene, we use DDColor~\citep{kang2022ddcolor} to produce RGB color of each image.
While high-quality colorization results can be obtained for each image using DDColor~\citep{kang2022ddcolor}, different image may be assigned with different color style, which makes it difficult to generate consistent 3D colorization results. However, \ourname{} successfully models the diverse results of colorization in the generated space, and each translated NeRF achieves a high degree of 3D consistency.
% \kunming{Also, please point out that this is the same chose as our baseline method.}
\end{itemize}

\subsection{Network Architecture}\label{sec: NeRF structure}
\begin{figure}[t]
\begin{center}
  \includegraphics[width=0.95\linewidth]{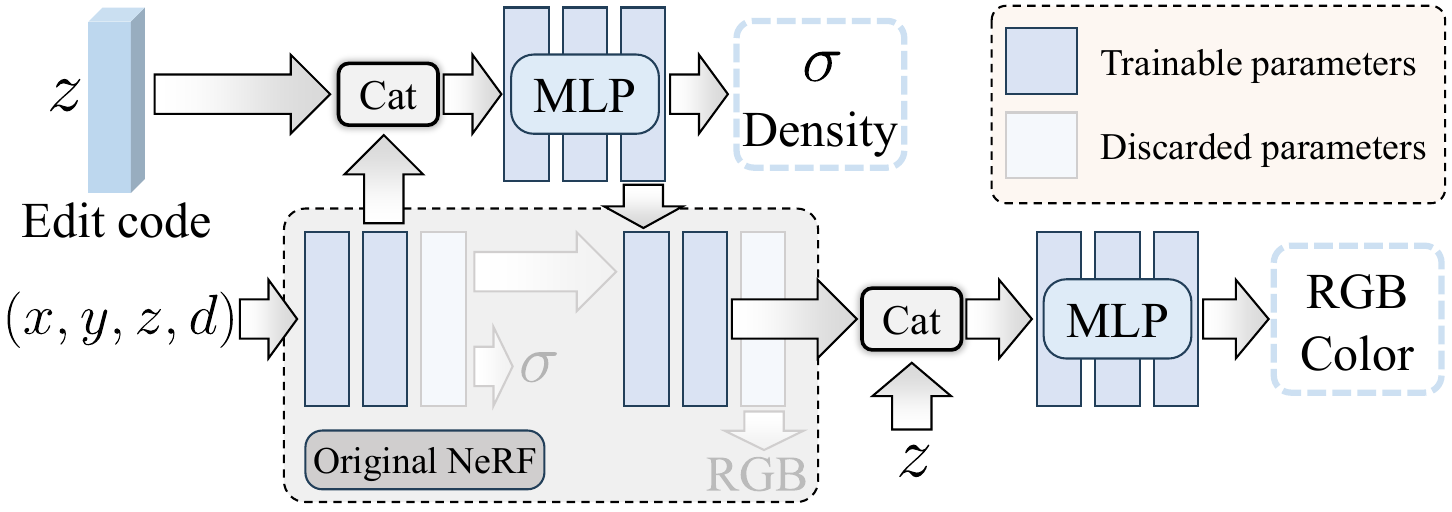}
\end{center}
   \caption{{\bf Detailed structure of the translated NeRF.} Given a pre-trained NeRF model, we concatenate our edit code $\mathbf{z}$ with its intermediate features and produce the density $\sigma$ and RGB color with two additional MLP networks. In this way, our translated NeRF is optimized to render the translated 3D scene conditioned on the edit code $\mathbf{z}$. } 
\label{fig:translated_NeRF}
\end{figure}

After 2D image editing, we then achieve NeRF editing in 3D domain by performing optimization of our translated NeRF using our well-designed loss functions. 
During the optimization process, we use our latent distil module to extract a latent code named edit code $\mathbf{z}$ from each edited 2D image.
Specifically, we employ the off-the-shelf VAE encoder from stable diffusion~\citep{rombach2022high} to extract the feature from the edited 2D image and then apply a tiny MLP network to produce this edit code $\mathbf{z} \in \mathbb{R}^{64}$. Just like a conventional variational encoder, this tiny MLP network is used to estimate the mean and variance of a Gaussian distribution and sample the edit code from it.
This tiny MLP network only contains three layers. 
% \kunming{xxx layers with xxx, describe the tiny MLP network.}
After extraction of the edit code $\mathbf{z}$, we then render novel views conditioned on the edit code using our translated NeRF.
In Fig~\ref{fig:translated_NeRF}, we provide the detailed structure of the translated NeRF. Our main purpose of this design is to make the translated NeRF render 3D scenes conditioned on the edit code $\mathbf{z}$. As such, the diversity of edits from 2D image-to-image translator can be modeled and represented by a Gaussian distribution of the edit code $\mathbf{z}$. 
% \kunming{(This sentence may not be accurate.)}
Given a pre-trained original NeRF, we discard its two layers used for density and color estimation. The edit code is concatenated with the intermediate features of the original NeRF and
and then fed into two additional MLP networks to obtain the density $\sigma$ and RGB color for volume rendering. During the optimization process of our \ourname{}, the original NeRF parameters except the discared parameters are updated, as well as the newly added MLP networks. 
% \kunming{Explain more and describe more in details. }

\section{Experiment Settings}

% \subsection{Evaluation Metrics}
% \kunming{We need to describe how we compute the metrics. Then we describe the datasets we used to evaluate our method. Finally, tell about the detailed implementation details. }

\subsection{Datasets and Evaluation Metrics}
Our \ourname{} is a unified NeRF-to-NeRF translation framework for various NeRF translation tasks such as text-driven NeRF editing, colorization, super-resolution, inpainting, etc. To verify the effectiveness of our \ourname, we conduct extensive experiments on various dataset and scenes to compare our \ourname{} with existing task-specific specialists, such as Instruct-NeRF2NeRF~\citep{haque2023instruct}, Palette-NeRF~\citep{kuang2023palettenerf}, NeRF-SR~\citep{wang2022nerf} and SPIn-NeRF~\citep{mirzaei2023spin}.

For text-driven NeRF editing, we compare our method with existing methods Instruct-NeRF2NeRF~\citep{haque2023instruct} on portrait datasets the face dataset~\citep{haque2023instruct}, the Fangzhou self-portrait dataset~\citep{wang2023nerf} and the Farm and Campsite dataset~\citep{haque2023instruct}.
The Face dataset~\citep{haque2023instruct} comprises $65$ images capturing different views of a single person captured by a smartphone. The camera poses are extracted by using the PolyCam app. The Fangzhou self-portrait dataset~\citep{wang2023nerf} is collected from users utilizing a front-facing camera, resulting in a total of $100$ frames. The Farm and Campsite dataset~\citep{haque2023instruct} consists of outdoor 360-degree scenes captured by a camera, containing 250 frames in total, and we only use the former 100 frames for data efficiency. We choose metrics CLIP Text-Image Direction Similarity~\citep{haque2023instruct} and CLIP Direction Consistency~\citep{haque2023instruct} reported in Instruct-NeRF2NeRF to evaluate editing quality, coupled with Fréchet Inception Distance (FID)~\citep{heusel2017gans} to measure generative diversity. More specifically, the reference distribution used for calculating FID is the distribution of 2D edit images, such as InstructPix2Pix~\citep{brooks2022instructpix2pix} editing results in the text-driven editing task. We employ FID to assess how closely our generated results align with the reference distribution.

For colorization, the LLFF dataset~\citep{mildenhall2021nerf} for quantitative comparison consists of three large-scale outdoor scenes and five indoor scenes. We also select part of the BlendedMVS dataset~\citep{yao2020blendedmvs} for more qulitative results, which covers a variety of scenarios, including cities, buildings, sculptures, and small objects. Following 2D colorization method~\citep{kang2022ddcolor}, we use colorfulness score (CF)~\citep{larsson2016learning} to measure the richness of color in rgb form and vividness of colorized images.
% \kunming{This is copied from the previous version of ICLR. So please check if these dataset settings are correct. So many datasets are used for colorization, but in the experiment table, the dataset is not mentioned. so confusing, please make it clear. }

For NeRF super-resolution, we follow the existing methods NeRF-SR~\citep{wang2022nerf} to build high-resolution NeRF with training images down-scaled by $\times 4$.
We conduct the comparison with the same datasets, i.e., LLFF dataset~\citep{mildenhall2021nerf} mentioned previously and the Realistic Synthetic $360^\circ$ dataset~\citep{mildenhall2019local} containing $8$ synthetic objects with $100$ images. We employ the same metrics as NeRF-SR~\citep{wang2022nerf}: Peak Signal-to-Noise Ratio (PSNR), Structural Similarity Index Measure (SSIM) and Learned Perceptual Image Patch Similarity (LPIPS).
% \kunming{add details and then describe how we conduct our experiments on the dataset. please point out that our training-testing prototext is the same as previous methods so the comparison is totally fair.}

For inpainting, existing methods SPIn-NeRF~\citep{mirzaei2023spin} and OR-NeRF~\citep{yin2023or} conducted comparative experiments on their customized dataset~\citep{mirzaei2023spin} of $10$ outdoor scenes, including $60$ training images with the object and $40$ test images without the object for each scene. For fair comparison, we follow these existing methods and also choose the same dataset as well as a statue dataset~\citep{mirzaei2023spin} to train our model and compute test metrics of Peak Signal-to-Noise Ratio (PSNR), Structural Similarity Index Measure (SSIM) and Learned Perceptual Image Patch Similarity (LPIPS) on the same test views as previous methods. 
% \kunming{re-organize and add more details}

\begin{figure*}[ht]
\begin{center}
  \includegraphics[width=0.95\linewidth]{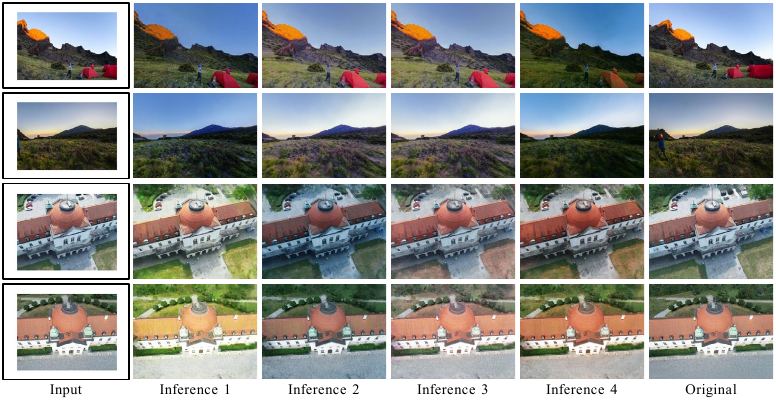}
\end{center}
   \caption{{\bf Application of our \ourname{} for NeRF zoom out.} We crop the original scene (right) as the input (left) scene and perform zoom out by our \ourname{}. We show our rendered views with different edit codes (inference $1- 4$) in our inference stage.
   As can be seen, our \ourname{} can acheive NeRF zoom out application so as to enlarge the original input 3D scene by generating plausible content with 3D geometry consistency. } 
\label{fig:application_zoomout}
\end{figure*}

\subsection{Implementation Details}
In our pipeline, we first perform 2D image-to-image translation and then lift those 2D edits up to 3D domain using our well-designed framework. Note that in our \ourname, we can flexibly choose a different plug-and-play 2D editor to support various NeRF editing task.
% \kunming{Is it true in this version now?}.

For text-driven NeRF editing, we leverage InstructPix2Pix~\citep{brooks2022instructpix2pix} as the 2D image-to-image translator to generate per-frame edited images corresponding to the unified text prompt. During this step, we follow Instruct-NeRF2NeRF~\citep{haque2023instruct} to randomly select the image similarity degree $S_{I}$ from $\{0.5, 2.5\}$, and text similarity degree $S_{T}$ from $\{6.0, 8.5\}$, producing edited images with significant diversities under the same text prompt. 
% \kunming{say what is image similarity degree and text similarity degree, at least add citations. no need} 
% Finally, $28$ stylized scene images are generated from the original multi-view images. {not 28!!! }\kunming{maybe this is the reason we have more diversity than I-N2N, maybe reviewers will ask, so if I-N2N is learned on the same edited images, what will happen? So if we do want to write down those details, we should ensure that our baseline is performed on the same way. }

After 2D editing, we then perform NeRF editing using our well-designed models and loss functions.
We implement all our \ourname{} based on PyTorch. 
Follow Instruct-NeRF2NeRF~\citep{haque2023instruct}, we use the original NeRF trained by using NeRFStudio~\citep{nerfstudio} on the original scene. Then we modify the original NeRF model as our translated NeRF as described in Sec.~\ref{sec: NeRF structure}.
During the training phase, we efficiently sample one image per iteration and extract $16,384$ rays with $48$ points per ray in a batch. 
% \kunming{Describe more about how we compute loss functions? Are there some hyperparameters for loss functions?}
Our model is trained using Adam optimizer with a learning rate of $1e-2$, running for $10,000-20,000$ iterations per scene. The total training phase takes about $3-8$ hours on one NVIDIA V100 GPU. During the inference phase, we randomly sample $\mathbf{z}$ from a standard Gaussian distribution and render the generated edited NeRF from arbitrary viewpoints with corresponding style defined by the sampled $\mathbf{z}$. The inference time for a translated NeRF need around 250ms. And the rendered time for a 100-frame scene takes about 3 mins. 

Notice that, our method is comparable with Instruct-NeRF2NeRF~\citep{haque2023instruct} in training time, and is a lightweight feed-forward model during inference without any heavy components on one NVIDIA V100 GPU, as shown in Table~\ref{supp tabel:  comparision_table}. Note that Instruct-NeRF2NeRF~\citep{haque2023instruct} does not have the inference phrase and requires retraining every time to get a new result, and the diversity between different results is small. In contrast, our method can directly perform forward inference by sampling different style codes to generate diverse results.

% \vspace{-8pt}
\begin{table}[h]
\centering
% \vspace{-8pt}
\resizebox{0.85\linewidth}{!}{
    \begin{tabular}{
            >{\arraybackslash}p{1.0cm}| % Method 
            >{\centering\arraybackslash}p{1.2cm} % time
            >{\centering\arraybackslash}p{1.2cm} % epoch
            >{\centering\arraybackslash}p{2.0cm}| % memory
            % >{\centering\arraybackslash}p{1.2cm} % time
            >{\centering\arraybackslash}p{1.2cm} % epoch
            >{\centering\arraybackslash}p{1.2cm} % memory
        }
        \hline
        \multirow{2}{*}{Method} & \multicolumn{3}{c|}{Train} & \multicolumn{2}{c}{Inference} \\
        \cline{2-6}
        & Time(h) & Iteration & Memory(GB) & FLOPs(G) & Latency(s)  \\
        \hline
        IN2N & 2.67 & 20000 & 18.32 & -- & -- \\
	    Ours  & 3.47 & 10000 & 20.92 & 131 & 0.35 \\
        \hline
    \end{tabular}
}
\caption{Comparsion with Instruct-NeRF2NeRF~\citep{haque2023instruct} on computational intensity.}
\label{supp tabel: comparision_table}
% \vspace{-12pt}
\end{table}

\section{More Insight Experiments}
\subsection{Quality of the Generation Space}
\begin{figure*}[ht]
\begin{center}
  \includegraphics[width=0.95\linewidth]{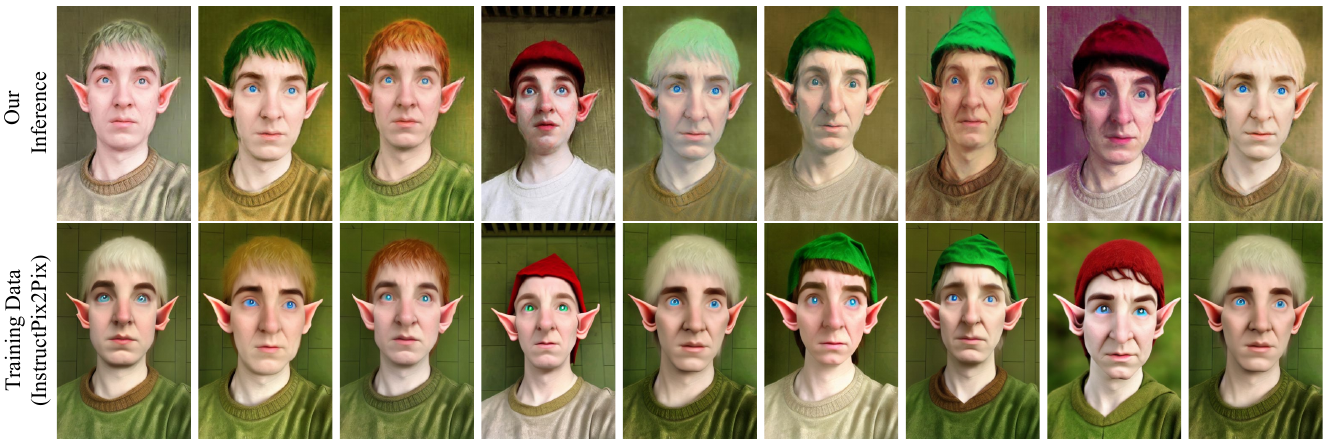}
\end{center}
   \caption{{\bf Image retrival results of our \ourname{} for Text-driven NeRF Editing.} We projected many of the generated results to all training viewpoints (top), and then performed image retrieval to find the closest match in the training data generated by InstructPix2Pix~\citep{brooks2022instructpix2pix} (bottom).} 
\label{fig:img_retrival}
\end{figure*}

To validate the quality of our generation space, we conducted an experiment as shown in Fig.~\ref{fig:img_retrival}, where we projected many of our generated results to all training viewpoints (top) and performed image retrieval to find the closest match in the training data (bottom). As illustrated in Fig.~\ref{fig:img_retrival}, many of our generated results are not present in the training data (InstructPix2Pix~\citep{brooks2022instructpix2pix}), demonstrating that our generation space is learned well.

\begin{figure}[t]
\begin{center}
  \includegraphics[width=0.95\linewidth]{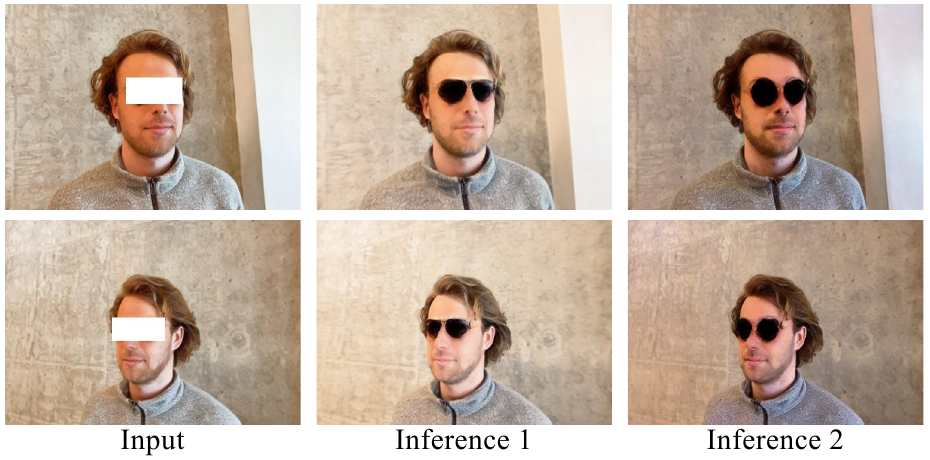}
\end{center}
   \caption{{\bf Application of our \ourname{} for text-driven inpainting.} We use ``sunglasses'' as the text condition to guide the inpainting process. } 
\label{fig:application_text_driven_inpainting}
\end{figure}

\subsection{Comparison with Naive Altering InstructPix2Pix Parameters in Instruct-NeRF2NeRF}
\begin{figure*}[t]
  \centering
   \includegraphics[width=0.95\linewidth]{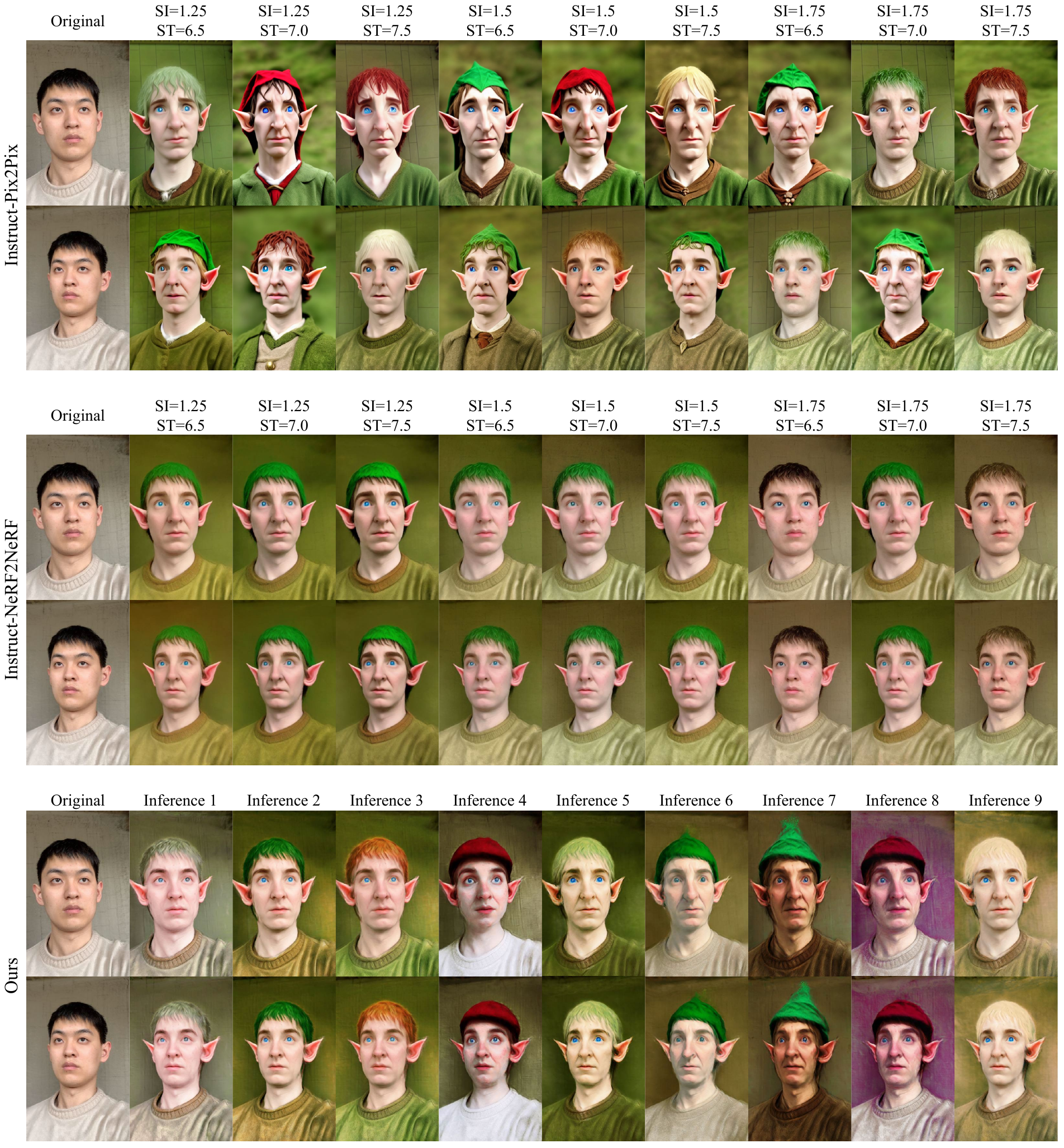}
   \caption{{\bf Diversity and view consistency comparisons of our \ourname{} with existing methods.} The same text instruction of ``Turn him into the Tolkien Elf'' is used for all the methods. As can be seen, InstructPix2Pix~\citep{brooks2022instructpix2pix} can produce diverse results in 2D but 3D multi-view consistency is not ensured. Though Instruct-NeRF2NeRF~\citep{haque2023instruct} can ensure the multi-view consistency, its results show little variance. In contrast, our \ourname{} can produce diverse editing results and address the 3D geometry consistency at the same time.
   }
   \label{fig:in2n_diff_elf}
\end{figure*}

We conducted this experiment as shown in Fig.~\ref{fig:in2n_diff_elf}, which demonstrates that altering the sampling parameters of the underlying 2D edit model cannot effectively increase diversity in Instruct-NeRF2NeRF~\citep{haque2023instruct}. Instruct-NeRF2NeRF~\citep{haque2023instruct} collapses on diversity due to two reasons. Firstly, the conditioning of InstructPix2Pix~\citep{brooks2022instructpix2pix} on the current NeRF rendering significantly collapses the diversity of InstructPix2Pix~\citep{brooks2022instructpix2pix}'s edit results, resulting in highly homogeneous editing outcomes. Secondly, Instruct-NeRF2NeRF~\citep{haque2023instruct} cannot ensure consistent edit directions during each iteration of update and edit, resulting in an average edit mode. 

\subsection{Interpolation of the Edit Code}
\begin{figure*}[t]
\begin{center}
  \includegraphics[width=0.85\linewidth]{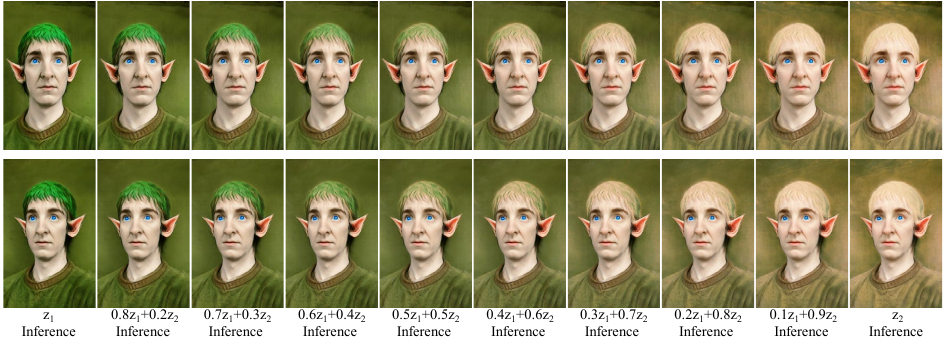}
\end{center}
   \caption{
{\bf Interpolation of the edit code.} We render the same views using the edit code $\mathbf{z}$ produced by interpolation of two random sampled edit code $\mathbf{z}_1$ and $\mathbf{z}_2$ by: $\mathbf{z} = \alpha*\mathbf{z}_1+(1-\alpha)*\mathbf{z}_2$. As can be seen, the rendering style of our translated NeRF model is highly related to the edit code and the style changes linearly when the interpolation weight $\alpha$ changes linearly. Note that, the 3D view consistency of the rendering scene is always maintained during the interpolation process. 
} 
\label{fig:interpolation}
\end{figure*}

We randomly sample two edit code $\mathbf{z}_1$ and $\mathbf{z}_2$ from Gaussian Distribution, and linearly interpolate nine latent code z by $\mathbf{z} = \alpha*\mathbf{z}_1+(1-\alpha)*\mathbf{z}_2$. Then we use these latent codes to directly inference the translated NeRFs. As shown in Fig.~\ref{fig:interpolation}, the rendering style of our translated NeRF model is highly related to the edit code and the style changes linearly when the interpolation weight $\alpha$ changes linearly. And the 3D view consistency of the rendering scene is always maintained during the interpolation process.

\begin{figure*}[t]
\begin{center}
  \includegraphics[width=0.95\linewidth]{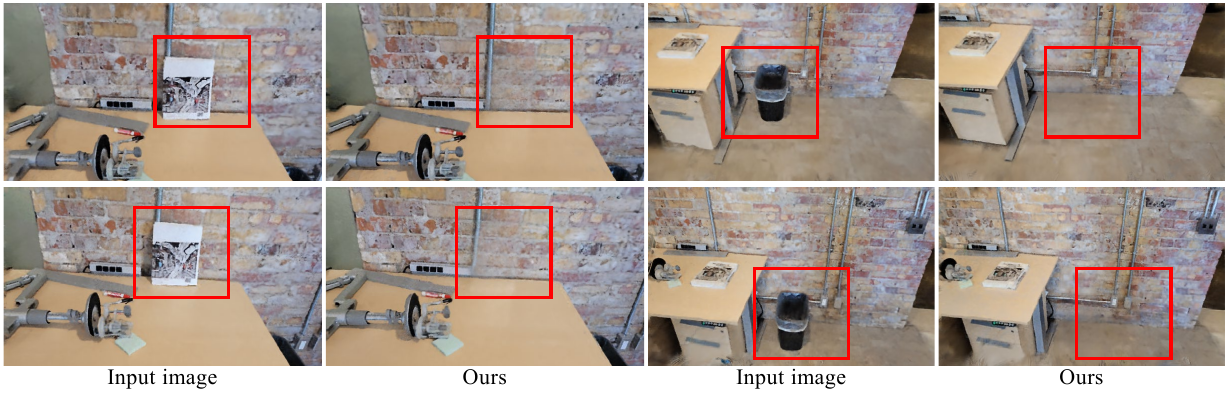}
\end{center}
   \caption{{\bf Qualitative results for NeRF inapinting.} We highlight the removed objects and our inpainting regions using the red boxes. } 
\label{fig:inpainting}
\end{figure*}

\begin{table}[t]
\centering
\resizebox{0.4 \textwidth}{!}{
    \begin{tabular}{
            >{\arraybackslash}p{0.8cm} % Method 
            >{\centering\arraybackslash}p{3.0cm} % text-driven editing
            >{\centering\arraybackslash}p{2.5cm} % text-driven editing
            >{\centering\arraybackslash}p{1.2cm} % text-driven editing
        }%
        \toprule
        \multirow{2}{*}{M} &CLIP Text-Image  & CLIP Direction & \multirow{2}{*}{FID $\downarrow$}  \\
        &Direction Similarity$\uparrow$  &Consistency $\uparrow$\\
        \midrule
       M=1 & 0.2635 & 0.9610 & 123.505   \\
        % \midrule
       M=3  & 0.2807  &  0.9650 & 91.823  \\
       M=5  & 0.2835 & 0.9638  &  86.377  \\
        \bottomrule
    \end{tabular}
}
    \caption{\textbf{Ablation of $M$.} }
\label{supp tabel: supp_M}
\end{table}

\subsection{Ablation on $M$}
The number of edits per perspective $M$ (default is 3) has little impact on the results. As each edit for every viewpoint is distinct, and we characterize this space by collectively utilizing all edits from different views (around 100). While a small $M$ value can have an impact on the overall data volume, once it reaches a certain threshold (\eg 3), the effect is negligible, as shown in Table~\ref{supp tabel: supp_M}.

\begin{table}[t]
\centering
\resizebox{0.3 \textwidth}{!}{
    \begin{tabular}{
            >{\centering\arraybackslash}p{3.5cm} % Method 
            >{\centering\arraybackslash}p{3.0cm} % CF
            % >{\centering\arraybackslash}p{3.0cm} % FID
        }%{|p{3cm}|p{3cm}|p{3cm}||p{3cm}|p{3cm}||p{3cm}|p{3cm}||p{3cm}|p{3cm}||p{3cm}|p{3cm}|}
        %\resizebox{\textwidth}{12mm}
        \toprule
        Method &CF $\uparrow$   \\
        \midrule
       PaletteNeRF\cite{kuang2023palettenerf} & 58.065     \\
        \midrule
       Ours & \textbf{75.960}    \\
        \bottomrule
    \end{tabular}
}
    \caption{\textbf{More quantitative results on colorization.} We compare our method with the state-of-the-art method PaletteNeRF~\citep{kuang2023palettenerf}. Since the latter does not provide an appropriate metric for comparison, we use colorfulness score (CF)~\citep{larsson2016learning} to measure the vividness of colorized images. We choose the dataset used by PaletteNeRF~\citep{kuang2023palettenerf}, namely, Fern, Horns, Flower and Orchids from the forward-facing LLFF dataset~\citep{mildenhall2021nerf}. }
\label{supp tabel: comparision_table_color}
\end{table}

\begin{table}[t]
\centering
\resizebox{0.5 \textwidth}{!}{
    \begin{tabular}{
            >{\centering\arraybackslash}p{3.2cm} % Method 
            >{\centering\arraybackslash}p{3.0cm} % text-driven editing
            >{\centering\arraybackslash}p{2.5cm} % text-driven editing
            >{\centering\arraybackslash}p{1.0cm} % text-driven editing
        }%
        \toprule
        \multirow{2}{*}{Method} &CLIP Text-Image  & CLIP Direction & \multirow{2}{*}{FID $\downarrow$}  \\
        &Direction Similarity$\uparrow$  &Consistency $\uparrow$\\
        \midrule
 
       Instruct-NeRF2NeRF         & 0.1383  & 0.9624  & 101.219   \\
        \midrule
       Ours  & \textbf{0.1583}  & \textbf{0.9683}  & \textbf{93.688}   \\
        \bottomrule
    \end{tabular}
}
    \caption{\textbf{More quantitative results on text-driven editing.} 
    We compare our method with the state-of-the-art method Instruct-NeRF2NeRF~\citep{haque2023instruct} with metrics reported in the latter. Following Instruct-NeRF2NeRF~\citep{haque2023instruct}, we conduct quantitative evaluation on bear dataset with 3 editing prompts and face dataset with 7 editing prompts.
    }
\label{supp tabel: comparision_table_editing}
\end{table}

\section{More Applications}
% Our \ourname{} can be support various NeRF translation tasks by choosing different 2D image-to-image translator. 
As we demonstrated before, we can achieve NeRF-to-NeRF text-drive 3D editing, super-resolution, colorization, and inpainting. To further demonstrate the versatility of our framework, we also provide two applications, NeRF-to-NeRF zoom out and text-driven inpainting, of our \ourname{} that have not been explored by previous methods.

\subsection{Zoom Out of NeRF}
Zoom out of NeRF is to extend an input NeRF along the input region to enlarge the NeRF scene.
% This is an unexplored problem that requires extrapolating off-camera 3D scene．
% \kunming{describe its difficulty and maybe why this task is not explored by previous methods, so as to emphasize that our framework is so powerful to support it. Then get into the details.}
Similar to inpainting, we also use LaMa~\citep{suvorov2022resolution} as the 2D image-to-image translator in our \ourname{} to solve the NeRF zoom out problem. Given source multi-view images of a 3D NeRF scene, we set the zoom out ratio as $1.25$ for image width and height to enlarge the source images. We first automatically generate binary masks for the zoom out region and then employ LaMa~\citep{suvorov2022resolution} to recover those zoom out regions. Since different content may be generated for zoom out regions in different 2D images from different viewpoint, it is difficult to ensure the 3D consistency in those zoom out regions. We show qualitative results of our zoom out application in Fig.~\ref{fig:application_zoomout}. As we can see that our method can successfully ensure the 3D consistency of those zoom out regions and generate reasonable NeRF scenes.

\subsection{Text-driven NeRF Inpainting}
% \kunming{What is the definition of text-driven NeRF inpainting, what's the difference between it and NeRF inpainting mentioned before. Please describe the difficulty of this task and maybe point out that why it is not explored by previous methods? Maybe because of the 3D inconsistency problem?} 
Text-driven NeRF Inpainting is similar to inpainting, but with the added restriction of text instructions. For the text-driven inpainting task, we use Blended Latent Diffusion~\citep{avrahami2023blended} as the 2D image-to-image translator, applying inpainting with text instructions. We get the mask in the same way mentioned in the inpainting task.
% We support various ways to get a mask, but note that multi-view masks must correspond to the same location in the 3D scene. For example, by artificially calculating the position of the part to paint in the 3D scene corresponding to the 2D image, or using the segment anything model~\citep{kirillov2023segment} to get the bounding box of the same object. 
Moreover, we found that if the text prompt is not provided as a guidance, the diffusion model tends to inpaint unreasonable content or monotonous colors close to the surroundings, instead of drawing meaningful objects. So we artificially set up the required text prompt or used the visual question answering model~\citep{gao2023llama} to get answers to the ``imagine what the white area might be" question.  For example, in this way, we can generate plausible 2D content in the mask area with the guidance of text instructions. After 2D editing, we then perform our proposed optimization to obtain the translated 3D NeRF scene.
% \kunming{Draw a conclusion of this implementation. } 
Qualitative results of our Text-driven NeRF inpainting results are shown in Fig.~\ref{fig:application_text_driven_inpainting}, where we can see that the area of mask is filled with content that matched the description of the text, such as various sunglasses.

\section{Qualitative Results Gallery}
\begin{figure*}[t]
\begin{center}
  \includegraphics[width=0.95\linewidth]{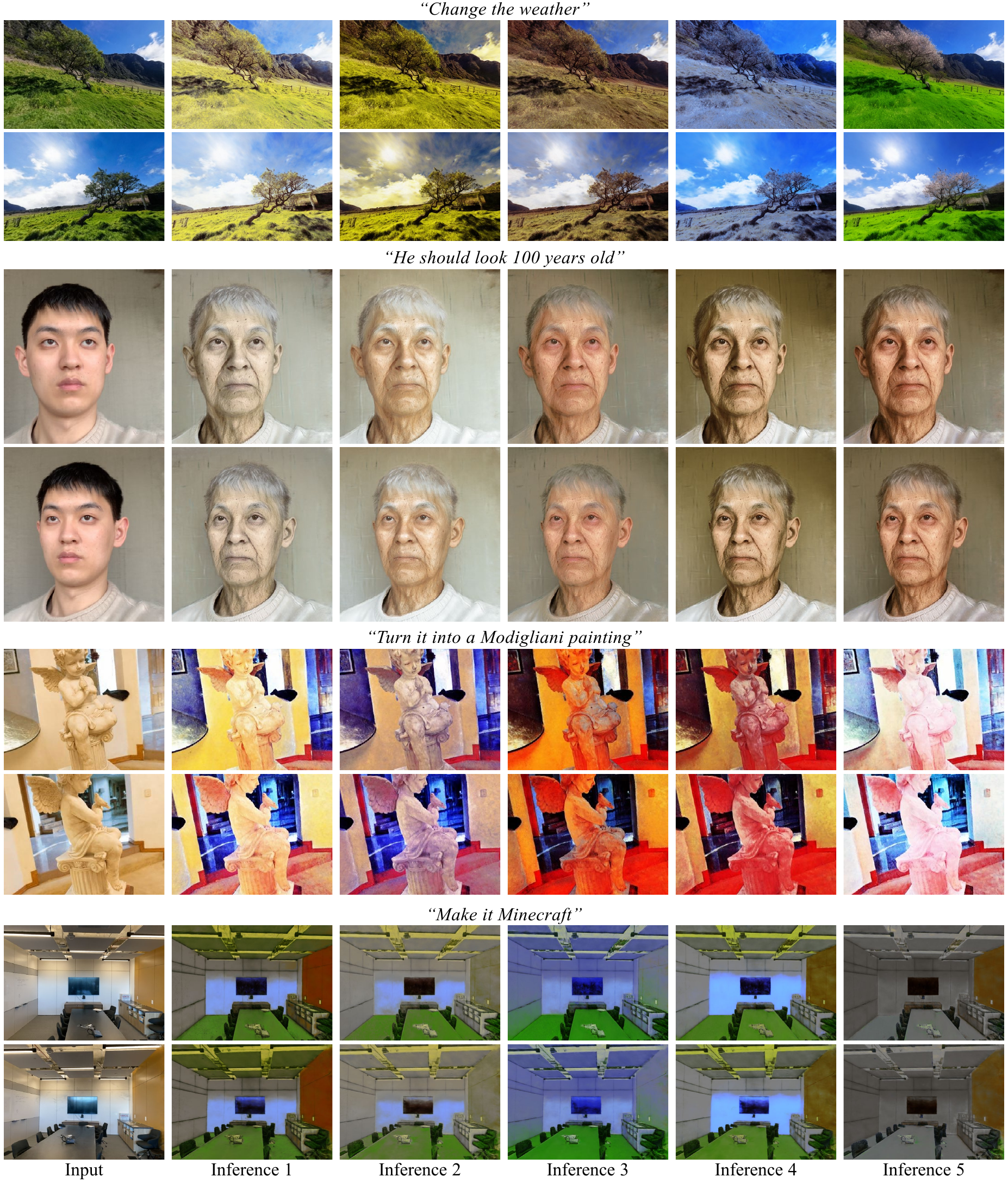}
\end{center}
   \caption{{\bf Qualitative results for Text-driven Editing.} We show the original 3D scene (left), our inference results (right) with different edit codes to show the diversity ability of our method. We can see that under different edit code, the edited scene with different styles can be rendered with plausible 3D geometry consistency. } 
\label{fig:Text-driven Editing}
\end{figure*}

\begin{figure*}[t]
\begin{center}
  \includegraphics[width=0.95\linewidth]{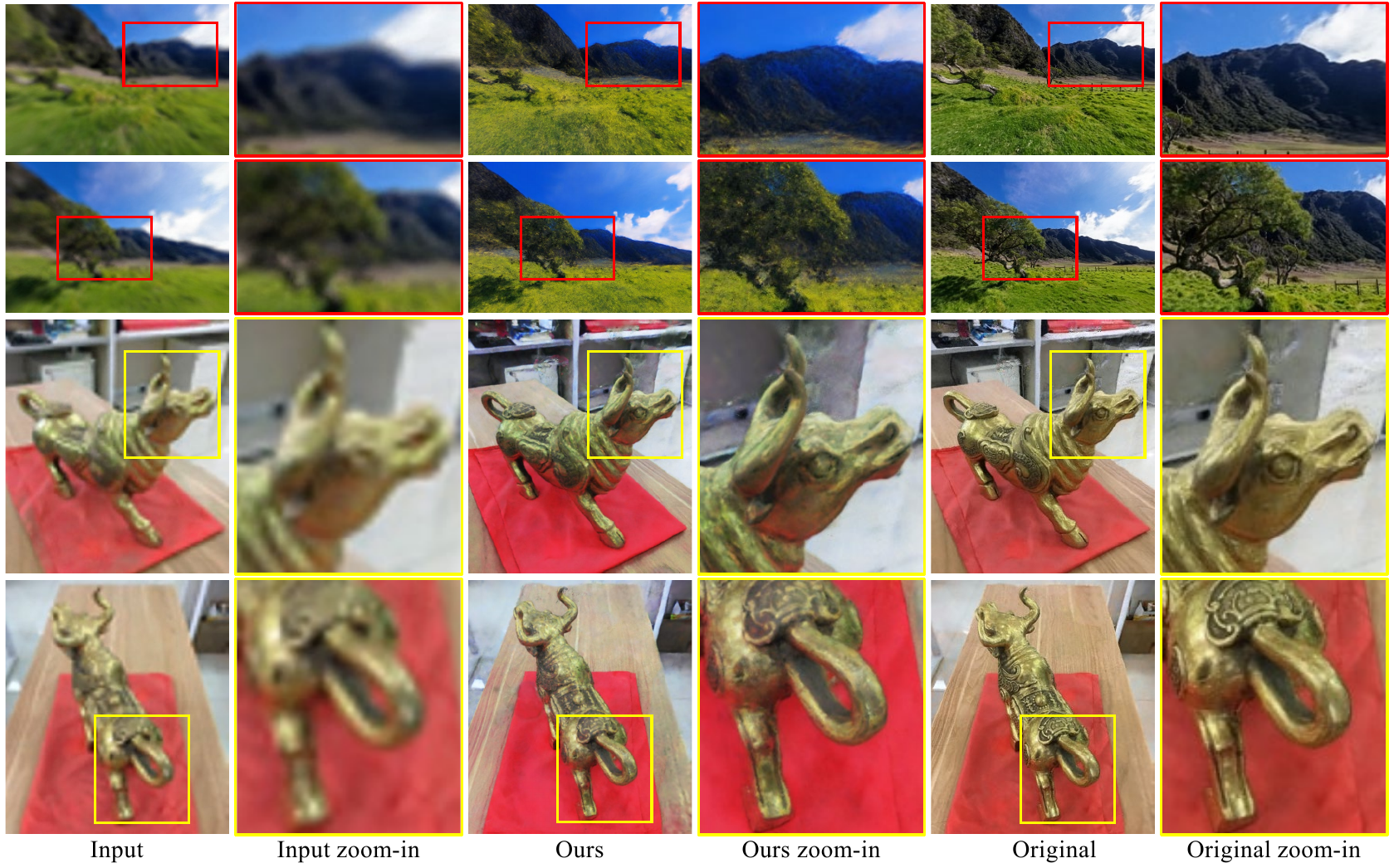}
\end{center}
   \caption{{\bf Qualitative results for NeRF super-resolution.} We show the original low-resolution input (left), our super-resolution result (middle) and the ground-truth (right) all with their zoom in results for better visualization. } 
\label{fig:sr}
\end{figure*}

\begin{figure*}[t]
\begin{center}
  \includegraphics[width=0.95\linewidth]{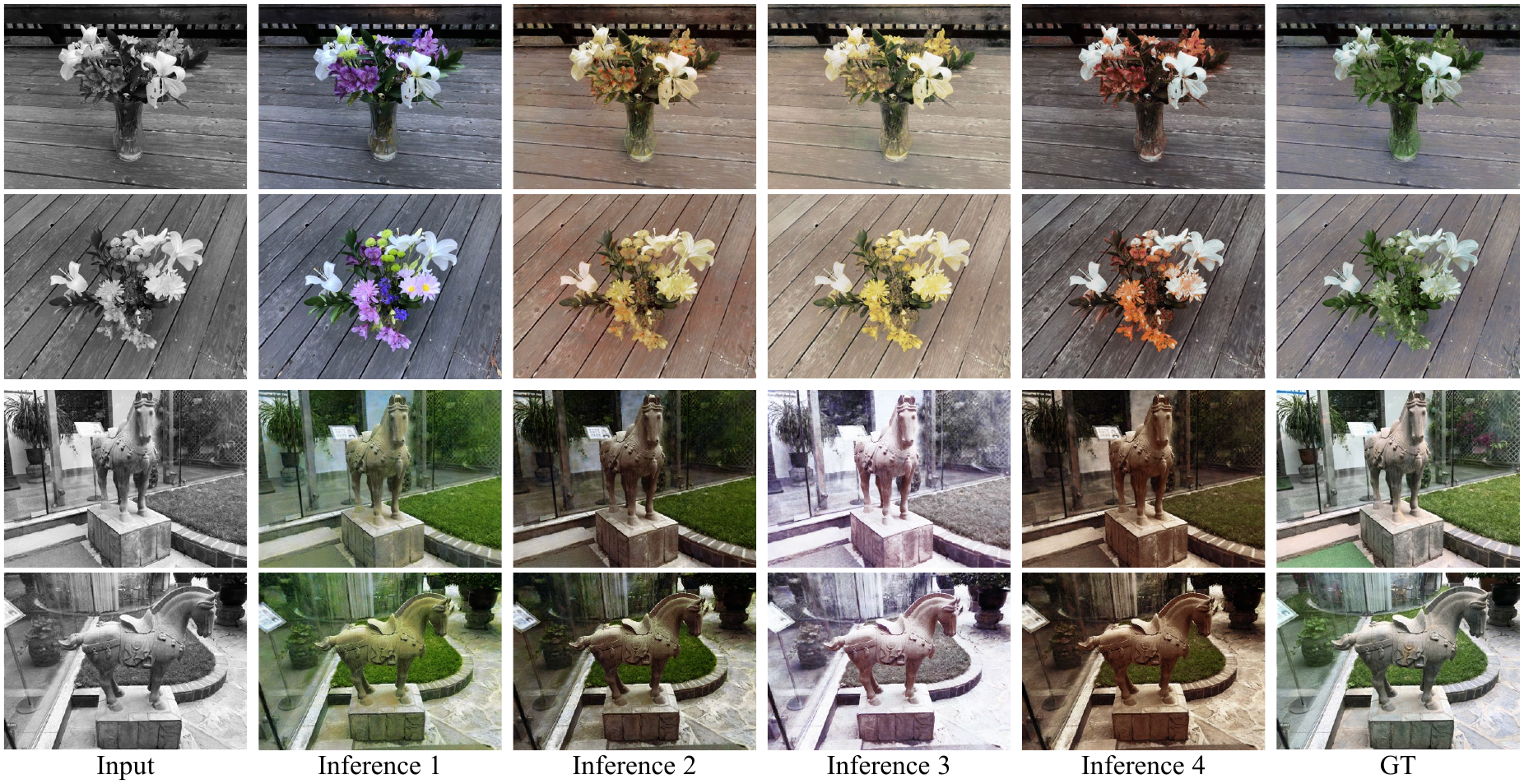}
\end{center}
   \caption{{\bf Qualitative results for NeRF colorization.} For each scene, we show two views of
the input gray-scale scene (left), five of our inference edited NeRF rendering results from different edit code (middle), and
the Ground-truth scene (right). As can be seen, our \ourname{} can produce plausible colorization results while maintaining the 3D multi-view consistency.  } 
\label{fig:supp color}
\end{figure*}

We provide more qualitative in Fig.~\ref{fig:inpainting}, Fig.~\ref{fig:Text-driven Editing}, Fig.~\ref{fig:sr}, and Fig.~\ref{fig:supp color}. For better visualization, we refer the reader to our project page:  \url{https://xiangyueliu.github.io/GenN2N/}.

% \subsection{Text-driven Editing}

%%%%%%%%%%%%%%%%%%%%%%%%%%%
\section{More Quantity Results}
We provide more qualitative in Table~\ref{supp tabel: comparision_table_editing} and Table~\ref{supp tabel: comparision_table_color}. For better visualization, we refer the reader to our project page.

\end{document}